%% file: main.tex
\documentclass[acmtog]{acmart}
\usepackage{lipsum}
\usepackage{xcolor}
\usepackage{bbding}
\usepackage{bm}
\usepackage{enumitem}
\usepackage{multirow}
\usepackage{xspace}

\usepackage{listings}
\usepackage{xcolor}
\usepackage{stfloats}

\newcommand{\methodname}{CAST\xspace}

\AtBeginDocument{%
  }

\acmSubmissionID{364}

\begin{document}

\def\name{CAST: Component-Aligned 3D Scene Reconstruction from an RGB Image}

\title{\name}

\author{Kaixin Yao}
\authornote{Equal contributions.}
\orcid{0009-0005-2056-6057}
\affiliation{%
    \institution{ShanghaiTech University}
    \city{Shanghai}
    \country{China}
}
\affiliation{%
    \institution{Deemos Technology Co., Ltd.}
    \city{Shanghai}
    \country{China}
}
\email{yaokx2023@shanghaitech.edu.cn}

\author{Longwen Zhang}
\authornotemark[1]
\orcid{0000-0001-8508-3359}
\affiliation{%
    \institution{ShanghaiTech University}
    \city{Shanghai}
    \country{China}
}
\affiliation{%
    \institution{Deemos Technology Co., Ltd.}
    \city{Shanghai}
    \country{China}
}
\email{zhanglw2@shanghaitech.edu.cn}

\author{Xinhao Yan}
\orcid{0009-0003-4736-5664}
\affiliation{%
    \institution{ShanghaiTech University}
    \city{Shanghai}
    \country{China}
}
\affiliation{%
    \institution{Deemos Technology Co., Ltd.}
    \city{Shanghai}
    \country{China}
}
\email{yanxh@shanghaitech.edu.cn}

\author{Yan Zeng}
\orcid{0009-0007-4612-8789}
\affiliation{%
    \institution{ShanghaiTech University}
    \city{Shanghai}
    \country{China}
}
\affiliation{%
    \institution{Deemos Technology Co., Ltd.}
    \city{Shanghai}
    \country{China}
}
\email{zengyan2024@shanghaitech.edu.cn}

\author{Qixuan Zhang}
\authornote{Project leader.}
\orcid{0000-0002-4837-7152}
\affiliation{%
    \institution{ShanghaiTech University}
    \city{Shanghai}
    \country{China}
}
\affiliation{%
    \institution{Deemos Technology Co., Ltd.}
    \city{Shanghai}
    \country{China}
}
\email{zhangqx1@shanghaitech.edu.cn}

\author{Wei Yang}
\orcid{0000-0002-1189-1254}
\affiliation{%
    \institution{Huazhong University of Science and Technology}
    \country{China}
    \city{Shanghai}
}
\email{weiyangcs@hust.edu.cn}

\author{Lan Xu}
\authornote{Corresponding author.}
\orcid{0000-0002-8807-7787}
\affiliation{%
    \institution{ShanghaiTech University}
    \country{China}
    \city{Shanghai}
}
\email{xulan1@shanghaitech.edu.cn}

\author{Jiayuan Gu}
\authornotemark[3]
\orcid{0000-0002-3207-7921}
\affiliation{%
    \institution{ShanghaiTech University}
    \country{China}
    \city{Shanghai}
}
\email{gujy1@shanghaitech.edu.cn}

\author{Jingyi Yu}
\authornotemark[3]
\orcid{0000-0002-8580-0036}
\affiliation{%
    \institution{ShanghaiTech University}
    \country{China}
    \city{Shanghai}
}
\email{yujingyi@shanghaitech.edu.cn}


\begin{abstract}
\label{sec:abstract}
    \input{sec/0_abstract.tex}

\end{abstract}



\begin{teaserfigure}
    \centering
  \includegraphics[width=\textwidth]{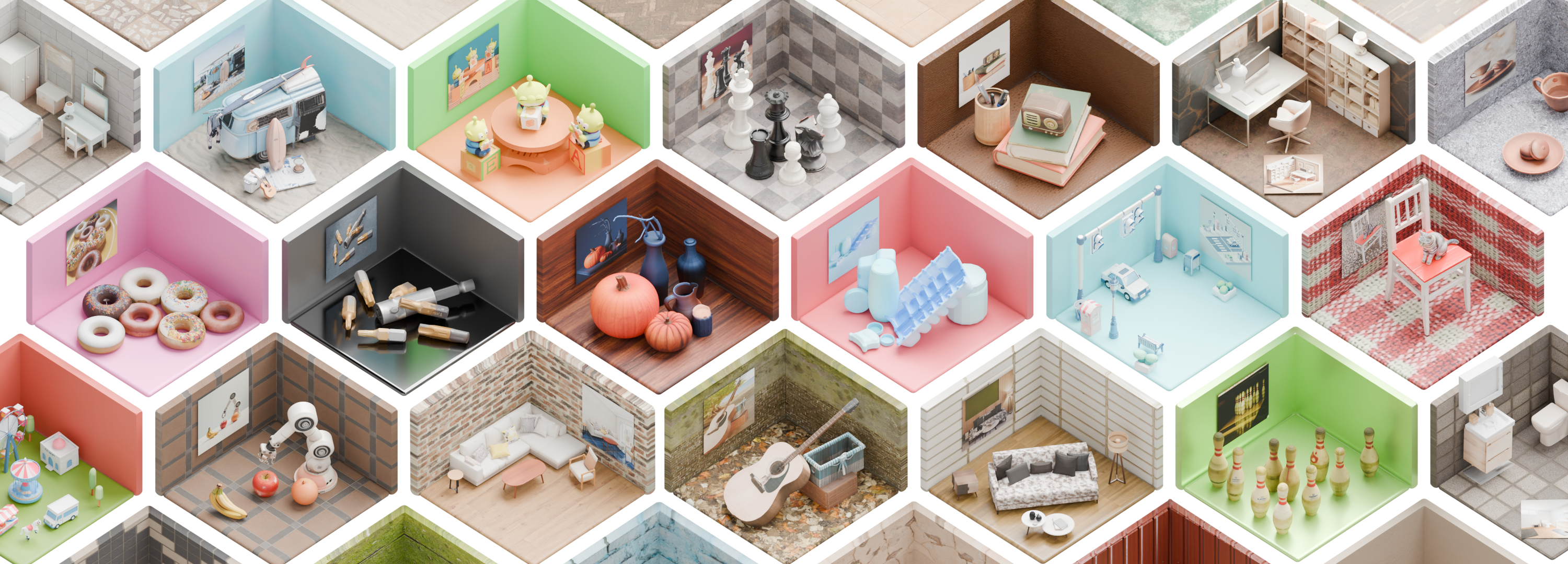}
  \caption{CAST brings diverse 3D scenes to life from a single image, where the relations between objects shaped by their physical roles and interactions come together to form a cohesive and immersive virtual environment.}
  \label{fig:teaser}
\end{teaserfigure}



\maketitle

\section{Introduction}
\label{sec:intro}
    \input{sec/1_intro.tex}

\section{Related Work}
\label{sec:related}

\input{sec/2_related.tex}

    
\input{sec/3_overview_part1.tex}

\label{sec:method3}

\input{sec/4_method_part2.tex}
\label{sec:method4}

\input{sec/5_method_part3_jigu.tex}
\input{sec/6_exp.tex}

\section{Conclusions}
\label{sec:conclusions}
    \input{sec/7_conclusion.tex}


\begin{acks}
This work was supported by National Key R\&D Program of China (2022YFF0902301), NSFC programs (61976138, 61977047), STCSM (2015F0203-000-06), and SHMEC (2019-01-07-00-01-E00003). We also acknowledge support from Shanghai Frontiers Science Center of Human-centered Artificial Intelligence (ShangHAI), MoE Key Lab of Intelligent Perception and Human-Machine Collaboration (ShanghaiTech University), Core Facility Platform of Computer Science and Communication of ShanghaiTech University, and HPC Platform of ShanghaiTech University.
\end{acks}

\bibliographystyle{ACM-Reference-Format}
\bibliography{main}

\newpage
\ 
\newpage
\appendix
\input{sec/appendix}




\end{document}

%% file: sec/0_abstract.tex
Recovering high-quality 3D scenes from a single RGB image is a challenging task in computer graphics. Current methods often struggle with domain-specific limitations or low-quality object generation. To address these, we propose CAST (Component-Aligned 3D Scene Reconstruction from a Single RGB Image), a novel method for 3D scene reconstruction. CAST starts by extracting object-level 2D segmentation and relative depth information from the input image, followed by using a GPT-based model to analyze inter-object spatial relations. This enables understanding of how objects relate to each other within the scene, ensuring more coherent reconstruction. CAST then employs an occlusion-aware large-scale 3D generation model to independently generate each object's full geometry, using Masked Auto Encoder (MAE) and point cloud conditioning to mitigate the effects of occlusions and partial object information, ensuring accurate alignment with the source image's geometry and texture. To align each object with the scene, the alignment generation model computes the necessary transformations, allowing the generated meshes to be accurately placed and integrated into the scene's point cloud. Finally, CAST applies a physics-aware correction mechanism, which leverages a fine-grained relation graph to generate a constraint graph. This graph guides the optimization of object poses, ensuring physical consistency and spatial coherence. By utilizing Signed Distance Fields (SDF), the model effectively addresses issues such as occlusions, object penetration, and floating objects, ensuring that the generated scene accurately reflects real-world physical interactions. Experimental results demonstrate that CAST significantly improves the quality of single-image 3D scene reconstruction, offering enhanced realism and accuracy in scene understanding and reconstruction tasks. CAST has practical applications in virtual content creation, such as immersive game environments and film production, where real-world setups can be seamlessly integrated into virtual landscapes. Additionally, CAST can be leveraged in robotics, enabling efficient real-to-simulation workflows and providing realistic, scalable simulation environments for robotic systems.

%% file: sec/1_intro.tex
Humans exist within clear networks of relations—family, friends, coworkers—that guide our decisions and behaviors. These connections shape our world and give it structure. Similarly, objects in a space also function within their own networks~\cite{latour2005reassembling}, but less noticed. They do not just exist in isolation; their placement, design, and material arise from physical constraints, functional roles, and human design intentions and influence how we move, interact, and perceive space. For example, a chair leans against a table for support, a cup rests on a saucer, and a lamp’s light interacts with surrounding surfaces, casting shadows that shape the overall scene. Recognizing these relations is critical for accurate scene parsing, modeling, and, more recently, 3D generation, ensuring virtual environments feel as realistic and coherent as the real world.

Significant progress has been made in generating single objects from text or image prompts. Neural rendering approaches~\cite{poole2022dreamfusion,wang2024prolificdreamer} optimize implicit representations, while native 3D generators~\cite{zhang20233dshape2vecset,zhang2024clay,xiang2024structured} directly create 3D shapes and textures via end-to-end learning. While these methods show promise for individual objects, applying them to generate entire scenes by assembling objects sequentially faces with notable shortcomings. A key challenge is accurate pose estimation. Existing methods often assume objects are view-aligned, which is rarely the case in real-world scenes. Objects may appear in diverse orientations, constrained by design, physics, or partial occlusion. Yet, most existing methods prioritize geometric fidelity over pose alignment, leaving this critical aspect underexplored.

An even more fundamental issue arises from the lack of inter-object spatial relations. Even with accurate poses, generated scenes often suffer from physically implausible artifacts: objects penetrate one another, float, or fail to make contact where necessary. These errors stem from the absence of spatial and physical constraints that naturally bind objects together, much as human relations structure our social world. While some recent methods~\cite{liu2022towards,zhang2024improving} encode spatial relations implicitly using encoder-decoder architectures, they remain limited to specific domains such as indoor scenes. Other scene-level generators~\cite{dogaru2024generalizable} position objects in a global coordinate system but neglect their relative poses and dependencies, further compromising realism and usability for downstream applications like editing, animation, and simulation.

To this end, we propose \methodname, a \emph{C}omponent-\emph{A}ligned 3D \emph{S}cene reconstruction method for compositional reconstruction of a 3D scene from a single RGB image. \methodname generates high-quality 3D meshes for individual objects, along with their similarity transformations (rotation, translation, scale), ensuring alignment with the reference image and enforcing physically sound interdependencies. \methodname starts by processing an unstructured RGB image using 2D foundation models (e.g., Florence-2~\cite{xiao2024florence}, GroundingDINO~\cite{liu2025grounding}, SAM~\cite{ravi2024sam}, Grounded-SAM~\cite{ren2024grounded}) to recognize, localize, and segment objects in an open-vocabulary manner. Off-the-shelf monocular depth estimators~\cite{wang2024moge} provide partial 3D point clouds and initial estimates of inter-object spatial relations, including relative transformations and scales.

The first core component of \methodname is our perceptive 3D instance generator with two modules: an occlusion-aware object generation module and a pose alignment generation module. The object generation module employs a latent diffusion-based generative model to produce high-fidelity object meshes conditioned on partial image segments and optional point clouds. This module incorporates an occlusion-aware 2D image encoder capable of inferring occluded regions, ensuring robust feature extraction for image conditions. To improve robustness to real-world point cloud conditioning, we simulate partial point clouds with occluded regions during training, enabling the model to handle occlusion effectively. 
The pose alignment module features an alignment generative model that produces a transformed partial point cloud, aligning with the complete geometry implicitly represented in the latent space. The similarity transformation is derived from the generated transformed point cloud and the partial point cloud estimated from the camera. Unlike direct pose regression methods~\cite{kehl2017ssd,labbe2020cosypose}, our method estimates transformations through generation, capturing the multi-modal nature of pose alignment.

The second core component of \methodname addresses inter-object spatial relations. Despite accurate pixel alignment, physically implausible artifacts such as penetration or floating can occur without explicit modeling of physical constraints. \methodname introduces a physics-aware correction process to ensure spatial and physical coherence. GPT-4v~\cite{achiam2023gpt} is utilized to identifies commonsense physical relations grounded in the input image, which are then used to optimize object poses based on these constraints. This process ensures that reconstructed scenes exhibit realistic physical interdependencies, making them suitable for applications like simulation, editing, and rendering.

Remarkably, \methodname excels at generating perceptually realistic 3D scenes from a wide range of images, whether they are sourced from indoor or outdoor settings, real-world captured, or AI-generated. Unlike previous approaches~\cite{liu2022towards,dai2024automated}, \methodname supports open-vocabulary reconstruction, even for challenging, in-the-wild images, thanks to our deliberate pipeline design.
Quantitatively, \methodname surpasses strong baselines in the indoor dataset, 3D-Front~\cite{fu20213d}, regarding object- and scene-level geometry quality. It also outperforms on perceptual and physical realism across a diverse set of images, including in-the-wild scenarios, as verified by visual-language models and user studies.

Given only a single image, \methodname can faithfully reconstruct the scene, with detailed geometry, vivid textures of the objects, and more importantly, the spatial and physical interdependencies between them. 
This capability democratizes virtual creation: a single snapshot of a room or outdoor space becomes a fully realized 3D environment, where objects are precisely posed, interact naturally, and account for occlusions. Game developers can integrate real-world setups into immersive landscapes, and filmmakers can effortlessly generate intricate virtual sets—unlocking creative potentials.
Beyond entertainment, \methodname paves the way for smarter robots. It can facilitate the real-to-simulation pipeline~\cite{li2024evaluating,torne2024reconciling} by enabling robotics researchers to construct digital replicas from real-world demonstration datasets with more efficient and scalable simulation workflows.

%% file: sec/2_related.tex
%
Transforming real-world scenes into the digital realm enhances our ability to understand, recreate, and interact with the 3D world around us. This practice is widely embraced in industries such as animation, film, gaming, architecture, and manufacturing. It enables the creation of immersive movie experiences, the digital preservation of historical relics, and the development of interactive environments for gaming.
For example, James Cameron employed groundbreaking 3D scanning technology in \textit{Avatar} (2009) to bring the lush, realistic world of Pandora to life. Similarly, in the gaming industry, \textit{The Witcher 3: Wild Hunt} incorporated lifelike terrain and architectures inspired by real-world locations in Poland, blending authentic cultural and natural elements with imaginative, open-world exploration.

Photogrammetry is a widely used method to capture the physical world in high detail and translate it into digital form~\cite{MVS,chen2018deep,mildenhall2020nerf,barron2021mip,barron2022mip,muller2022instant,kerbl3Dgaussians}, but it requires tens to hundreds of images from multiple viewpoints, making it time-consuming, resource-intensive, and hard to scale. In contrast, single-image-based approaches are more efficient and scalable, requiring only one image that can be easily obtained from online repositories, eliminating the need for expensive scanning devices or multi-view setups.

\subsection{Single Image Scene Reconstruction}


Scene-level reconstruction from a single image presents challenges due to object diversity, occlusions, and the need to preserve spatial relations. A starting point is monocular depth estimation, where depth is inferred from a single image, typically generating a depth point cloud~\cite{yin2023metric3d,bhat2023zoedepth,piccinelli2024unidepth,wang2024moge,yang2024depth}. While it provides valuable information, it struggles with occlusions and hidden portions of the scene. 
To address this, novel view synthesis methods use representations like radiance fields~\cite{yu2021pixelnerf, tian2023mononerf, yu2022monosdf} and 3D Gaussian~\cite{szymanowicz2024splatter,szymanowicz2024flash3d}, learning occlusion priors from 3D datasets~\cite{geiger2013vision,dai2017scannet, chang2015shapenet,sun2018pix3d}. Despite these advances, monocular reconstruction methods still often struggle to provide detailed and precise scene representations.

Some methods focus on directly regressing geometries along with their semantic labels in the scene~\cite{dahnert2021panoptic, gkioxari2022learning, chu2023buol, chen2024single}. These approaches typically rely on scene datasets with ground truth object annotations, such as Matterport3D~\cite{fu20213d} and 3DFront~\cite{fu20213d}, which are often small in scale and limited to indoor room environments. However, the feed-forward nature of these methods leads to the generation of geometries that often lack sufficient detail and quality.


To better film real world to digital, other methods turn to retrieval-based approaches~\cite{langer2022sparc,gumeli2022roca,kuo2021patch2cad,gao2024diffcad,dai2024automated}, which enhance scene quality by searching for and replacing objects in a scene with similar objects from a pre-existing dataset. These methods incorporate advanced tools such as GPT-4~\cite{achiam2023gpt}, SAM~\cite{kirillov2023segment, ren2024grounded}, and depth priors to decompose scenes.
While these methods improve scene realism by integrating real-world objects, they are constrained by the richness and scope of the datasets they rely on. For scenes outside the dataset's domain, retrieval-based methods either produce erroneous results or fail to find suitable replacements, significantly degrading the quality of the reconstructed scene.

\begin{figure*} [ht]
  \centering
  \includegraphics[width=\textwidth]{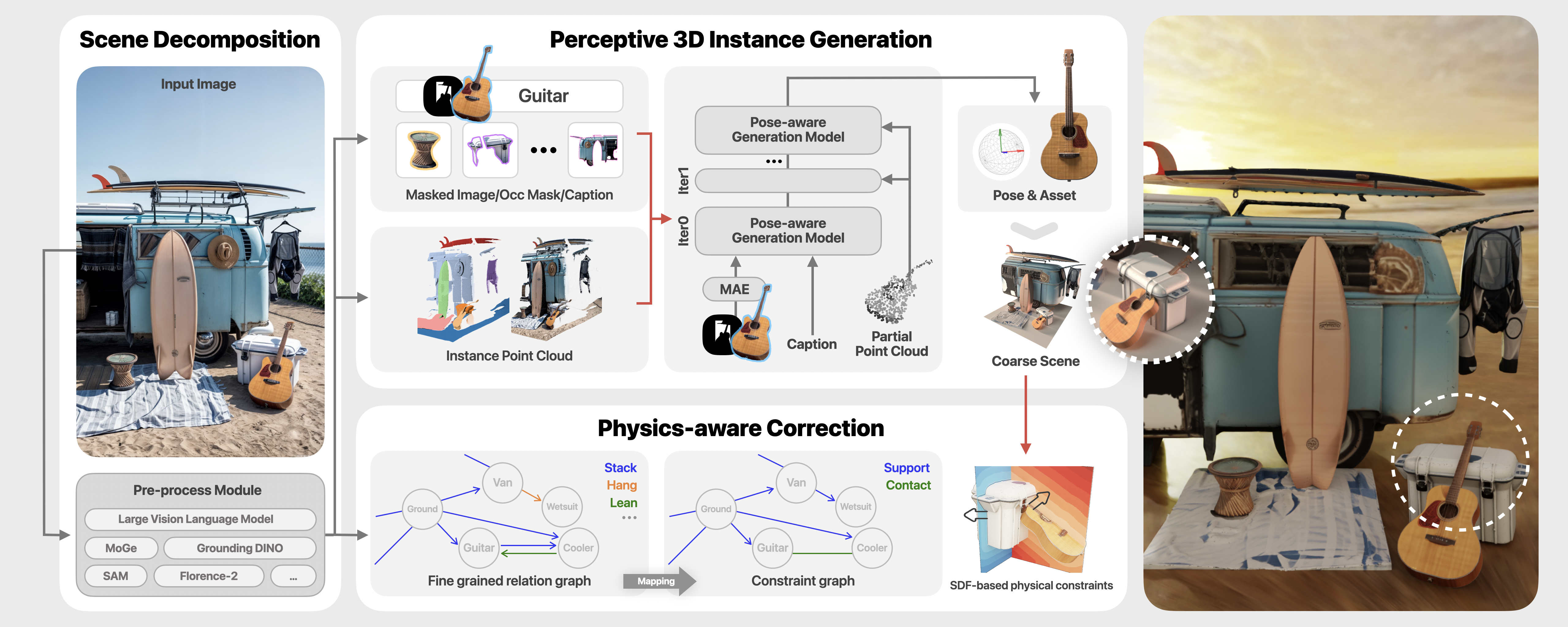}
  \captionof{figure}{
  Overview of the proposed pipeline. The input RGB image is processed through scene analysis to extract key information, followed by pose-aware generation to create initial 3D models. Physical constraint refinement ensures realistic interactions and spatial relations, yielding a high-quality, mesh-based 3D scene.
  } \label{fig_overview1}
\end{figure*}

%
\subsection{Reconstruction as Generation}
With the continuous advancements in the field, the ability to create high-quality 3D digital assets from various types of open-vocabulary images or text prompts has significantly improved. This advancement has precipitated a paradigm shift where the single-view reconstruction problem evolves into a generative 3D synthesis framework.
This paradigm change allows for the generation of 3D assets without being confined to a fixed dataset, enabling more flexible and scalable scene reconstruction.

%
Much of the current research in 3D asset generation focuses on distilling 3D geometry from 2D images generative models ~\cite{poole2022dreamfusion,tang2023dreamgaussian,wang2024prolificdreamer}.  More recent developments expand this approach by incorporating multi-view images for supervision~\cite{liu2023syncdreamer,liu2023zero,liu2024one,voleti2025sv3d,long2024wonder3d,wu2024unique3d}, often trained on rendered images from large-scale object datasets like Objaverse~\cite{deitke2023objaverse}, to enhance view consistency during generation.
Some approaches directly regress the shape and appearance of individual objects based on input image~\cite{hong2023lrm,tang2025lgm}. While these methods achieve satisfactory visual results, they frequently fail to reproduce fine geometric details.
To improve the quality of 3D geometry, a growing body of work has moved away from 2D supervision entirely, opting instead to train directly on 3D assets~\cite{deitke2023objaverse,deitke2024objaverse}. These methods produce high-quality object-level geometries with advanced processing techniques~\cite{zhang20233dshape2vecset, zhang2024clay, xiang2024structured}. However, such approaches focus on isolated objects and fail to address scene-level challenges, such as modeling spatial hierarchies, inter-object relations, and environment lighting.
%
Scene generation remains underdeveloped due to the high computational and representational complexity of modeling object relations, lighting, and materials. Despite progress, current approaches still struggle to produce fully realized, editable 3D scenes. Existing paradigms either use video diffusion models~\cite{ho2022video, ho2022imagen, blattmann2023stable} to generate navigable 2D projections~\cite{bruce2024genie, yu2024wonderjourney}, or rely on diffusion priors for volumetric scene approximations via 3D Gaussian splatting~\cite{wu2024reconfusion, gao2024cat3d, liang2024luciddreamer}. While these methods yield compelling visuals, they are incompatible with traditional production pipelines, lacking editable meshes, UV mappings, and decomposable PBR materials.

A more feasible paradigm decomposes scenes into modular components—objects, backgrounds, and environmental generating and reassembling them into an editable scene graph for greater flexibility and precision. For example, Gen3DSR~\cite{dogaru2024generalizable} uses DreamGaussian~\cite{tang2023dreamgaussian} for open-vocabulary reconstruction. However, it struggles with occlusions, pose estimation, and editing individual objects, while relying on 2D models leads to poor geometric details and low-fidelity representations.
Another recent work, Midi~\cite{huang2024midi}, learns spatial relations between objects in a scene but requires training on datasets with ground truth 3D meshes and annotations. This reliance on specific datasets limits its scalability and generalization to arbitrary scenes.

Our approach shares conceptual foundations with classical analysis-by-synthesis methods \cite{yuille2006vision}, as both aim to infer 3D structure by generating explanations for observed imagery. However, while analysis-by-synthesis relies on iterative rendering and pixel-level optimization, our method leverages pre-trained generative models and learned priors to synthesize plausible 3D scenes directly, often bypassing explicit rendering and optimization loops, thereby improving scalability, efficiency, and adaptability to open-world scenarios.

Building on this foundation, we present a novel scene reconstruction pipeline that generates each object independently and aligns them into a cohesive scene. Unlike existing methods, our approach preserves accurate geometry, textures, and consistent spatial relations, resulting in more realistic, reliable, and editable reconstructions with improved quality and flexibility.


\subsection{Physics-Aware 3D Modeling}
Generating physically plausible 3D assets is crucial for ensuring realism and functionality in applications such as animation, gaming, and robotics.
While recent 3D generative models excel at creating visually realistic objects, they often fall short in achieving physical plausibility. To address this limitation, physics-aware 3D generative models have been developed to integrate physical principles into the generation process. Some methods use soft-body simulation to animate 3D Gaussians~\cite{xie2024physgaussian,zhong2025reconstruction}, or generate articulated objects with physics-based penalties~\cite{liu2023few}, while others ensure self-supporting structures through rigid-body simulation~\cite{mezghanni2021physically,mezghanni2022physical,chen2024atlas3d} or FEM~\cite{guo2024physically,xu2024precise}. These methods leverage offline or online physical simulations to check the physical validity of generated shapes and in turn guide generation.
However, these approaches are typically confined to individual objects, overlooking the mutual influences between multiple objects within a scene. 

Incorporating physical constraints into scene synthesis is much more challenging due to the inclusion of more complex relations, e.g., inter-object contact. \citet{yang2024physcene} integrates constraints like object collisions, room layout, and object reachability into their scene-level generation pipeline. However, it is limited to indoor scene synthesis and relies on a closed-vocabulary database to perform shape retrieval.
\citet{ni2024phyrecon} addresses the issue of physical implausibility in multi-view neural reconstruction. It leverages both differentiable rendering and physical simulation to learn implicit representations. However, it requires multi-view images as input, focuses on individual objects, and primarily addresses only stability (simulating the dropping of objects).
In contrast, our method operates in an open-vocabulary setting and requires only a single input image. Furthermore, it accounts for more complex inter-object relations, particularly support and contact, making it more versatile and applicable across diverse scenarios.


%

%% file: sec/3_overview_part1.tex
\section{Overview}

Scene-level reconstruction from a single image is a fundamental challenge in computer graphics, with broad applications in animation, virtual reality, and interactive gaming. Unlike object-level reconstruction, which focuses on isolated objects, scene-level reconstruction emphasizes the arrangement and relations of multiple entities under realistic (or stylized) physics. By capturing per-object structures, spatial relations, and contextual cues, this holistic approach enables more immersive experiences, compelling narratives, and efficient workflows—benefits that surpass those of single-object reconstructions. Although previous methodologies have explored feed-forward pipelines or retrieval-based approaches using fixed 3D templates~\cite{liu2022towards,dai2024automated}, these methods often struggle to capture nuanced scene semantics and complex object relations. To address these limitations, we propose a generation-driven scene reconstruction approach with emphasized object relations to construct high-fidelity, contextually consistent 3D environments from a single, unannotated RGB image whether sourced from real-world photography or synthetic data (see Fig.~\ref{fig_overview1}).

A key insight of our method is the thorough object relation analysis of scene contextual information. First, we perform object segmentation to identify and localize constituent objects within the image. We then obtain preliminary geometric information, i.e., point clouds, and explore semantic and spatial relations among objects. This contextual backbone informs our subsequent object-wise generation pipeline, ensuring that each reconstructed object retains not only its geometric fidelity but also its correct placement in the broader scene. Finally, we synthesize a coherent 3D environment that respects physical plausibility—achieving structurally sound layouts and realistic interactions among scene elements.




Our research focuses on two primary objectives: to explore how generative models can effectively capture complex inter-object relations in order to produce realistic, scene-level reconstructions from a single image; and to identify strategies for integrating geometric cues and contextual information that maximize accuracy and plausibility in 3D reconstructions.
Through this investigation, we demonstrate that generative methods provide a more flexible and robust alternative to traditional feed-forward and retrieval-based techniques. These methods allow for fine-grained control over both object-level details and global scene composition, thus streamlining content creation pipelines for animation, game development, and other fields requiring accurate, visually compelling 3D models.
This work highlights the advantages of a generation-centric framework and lays the groundwork for future advancements in scene-level 3D reconstruction. It also underscores the growing importance of context-driven approaches in bridging the gap between 2D imagery and immersive, interactive virtual environments.

%
\paragraph{Preprocessing}
To facilitate comprehensive scene reconstruction from a single image, we first perform an extensive semantic extraction that provides a robust foundation for subsequent processing. Specifically, we employ Florence-2~\cite{xiao2024florence} to identify objects, generate their descriptions, and localize each object with bounding boxes. We then leverage GPT-4v~\cite{achiam2023gpt} to filter out spurious detections and isolate meaningful constituent objects, allowing for open-vocabulary object identification that is not constrained by predefined categories. Next, we use GroundedSAM-v2~\cite{ren2024grounded} to produce a refined segmentation mask $\{ \bm{M}_i \}$ for each labeled object $\{ \bm{o}_i \}$, thereby obtaining both precise object boundaries and corresponding occlusion masks, which play a crucial auxiliary role in the object generation stage.
Apart from semantic cues, we also integrate geometric information by extracting a scene-level point cloud. Using MoGe~\cite{wang2024moge}, we generate pixel-aligned point clouds $\{ \bm{q}_i \}$ for each object $\{ \bm{o}_i \}, i \in \{1,\dots,N\}$ and a global camera parameter in the scene coordinate system. This additional geometric data is subsequently matched to each object’s segmentation mask, providing a reliable structural reference for the final 3D scene reconstruction.

%% file: sec/4_method_part2.tex
\begin{figure} []
  \centering
  \includegraphics[width=0.98\linewidth]{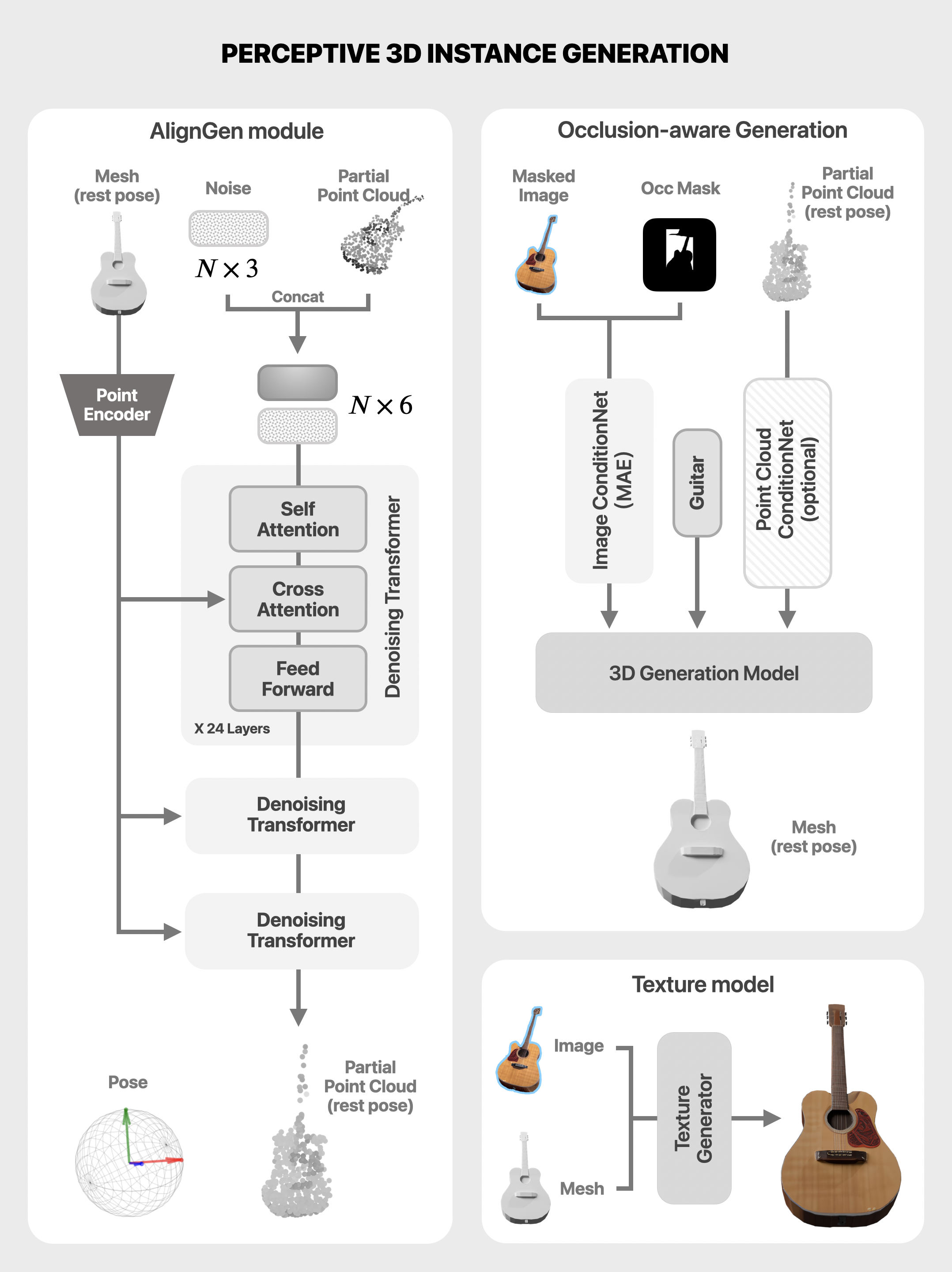}
  \captionof{figure}{Network design of our alignment generation model (Sec.~\ref{sec:transformationgen}), occlusion-aware object generation model (Sec.~\ref{sec:objectgen}), and an illustrative figure of the texture generation model.} \label{fig_overview2}
\end{figure}

\section{Perceptive 3D Instance Generation}
\label{sec:method2}

%

In the endeavor to reconstruct high-fidelity 3D scenes from single RGB images, a brute-force approach involves generating the entire scene mesh directly using techniques such as single-image depth estimation or diffusion priors. However, this method inherently struggles to manage occlusions, render invisible components, and accurately represent object relations due to the complex and intertwined nature of real-world scenes. 
Instead of generating an entire scene mesh directly, our approach focuses on individual object generation and then arranges the objects via precise relational alignment, as illustrated in~\ref{fig_overview2}. This strategy offers several advantages: 1. focusing on individual objects ensures higher geometric fidelity and allows for detailed modeling, resulting in more accurate and visually appealing scene components.
2. operating within a canonical space ensures that generated assets adhere to standardized orientations and scales, seamlessly integrating with artist-defined coordinate systems and promoting consistency across digital content creation tools.
3. the modular approach supports various applications such as editing, rendering, and simulation, enabling independent manipulation of objects for greater flexibility and efficiency.
By decomposing scene reconstruction into object-wise generation and alignment, our method improves asset quality and manageability while enhancing the overall coherence and functionality of the 3D environment. This approach addresses challenges like geometric precision and efficient post-processing, advancing single-image 3D scene generation.

Object-wise generation presents significant challenges, primarily due to partial observations of objects within a scene caused by occlusions and limited sensor coverage. Additionally, existing generation methods often fail to coordinate multiple objects cohesively, resulting in inconsistent and unrealistic scenes. To overcome these limitations, we propose an Occlusion-Aware 3D Object Generation framework that integrates partial observations with comprehensive scene understanding. 
Specifically, given an image and its point cloud, our framework generates a high-quality 3D asset that not only resembles the input image but also aligns accurately with the partial point cloud represented in its corresponding canonical space.
Furthermore, we compute a transformation matrix that maps the generated object from its canonical space back to the original scene space, ensuring spatial consistency within the scene.

A critical aspect of our object generation process is the utilization of a large generative model to generate holistic and high-fidelity object meshes from partial image and point cloud observations. To do so, we first follow state-of-art native 3D generative models~\cite{xiang2024structured,zhang20233dshape2vecset,zhang2024clay} to pre-train a large-scale 3D generative model conditioning on textual and image inputs. %

%
We build upon existing generative frameworks featuring 3DShape2VecSet representation~\cite{zhang20233dshape2vecset,zhang2024clay}, which prioritize geometry generation by utilizing a Geometry Variational Autoencoder (VAE). This VAE framework encodes uniformly sampled surface point clouds into unordered latent codes and decodes these latent representations into Signed Distance Fields (SDFs). Formally, the VAE encoder $\mathcal{E}$ and decoder $\mathcal{D}$ are defined as:
\begin{equation}
    \bm{Z}=\mathcal{E}(\bm{X}),\ \mathcal{D}(\bm{Z},\bm{p}) = \text{SDF}(\bm{p}),
\end{equation}
where $\bm{X}$ represents the sampled surface point cloud of the geometry, $\bm{Z}$ is the corresponding latent code, and $\text{SDF}(\bm{p})$  denotes the operation of querying the SDF value at point $\bm{p}$ for subsequent mesh extraction via marching cubes.
To effectively incorporate image information into the geometry generation process, we employ DINOv2~\cite{oquab2023dinov2} as our image encoder, following methodologies outlined in \citet{zhang20233dshape2vecset,zhang2024clay, xiang2024structured},
The geometry latent diffusion model (LDM) is then formulated as:
\begin{equation}
    \epsilon_\text{obj} (\bm{Z}_t; t,\bm{c}) \rightarrow \bm{Z},
\end{equation}
where $\epsilon$ represents the diffusion transformer model, $\bm{Z}_t$ is noisy geometry latent code at timestep $t$, and $\bm{c}$ denotes the encoded image features from DINOv2. We follow the pre-training process of prior works~\cite{zhang20233dshape2vecset,zhang2024clay} and pre-train the base model on Objaverse~\cite{deitke2023objaverse}. Upon training, our generation model $\epsilon$  is capable of generating detailed 3D geometry solely based on image features.

\subsection{Occlusion-aware 3D Object Generation}
\label{sec:objectgen}

Directly applying 3D generative-based models faces considerable challenges as
real-world scenarios often present challenges such as partial occlusions in the input images, which severely degrade the quality and accuracy of the generated object geometries. To address this issue, we leverage the Masked Auto Encoder (MAE) capabilities of DINOv2. Specifically, during inference, we provide an occlusion mask $\bm{M}$ alongside the input image $\bm{I}$, enabling the encoder to handle missing pixels by inferring latent features for the occluded regions. This is formalized as:
\begin{equation}
    \bm{c}_m=\mathcal{E}_\text{DINOv2}(\bm{I} \odot \bm{M}),    
\end{equation}
where $\bm{M}$ is a binary mask indicating which tokens should be masked and replaced with a [mask] token. During the pretraining phase, DINOv2 is trained with randomly set masks, allowing it to robustly infer missing parts based on the visible regions. Consequently, during inference, even if parts of the object image are occluded, the encoder can effectively reconstruct the necessary features, ensuring that the generative model maintains high-quality and accurate 3D reconstructions.
This integration of image conditioning and occlusion handling is pivotal for our pipeline, as it ensures that the generated 3D objects are both visually consistent with the input images and geometrically faithful to the underlying structure. 



\paragraph{Canonical Point Cloud Conditioning} 
Though our object generation model produces visually plausible meshes from input object images, it is challenging to generate pixel-aligned geometry due to the high-level nature of the encoded image condition $\bm{c}$ and the absence of pixel-wise supervision. We address this issue by additionally conditioning our object generation model on observed partial point clouds in canonical coordinates.
This dual conditioning ensures that the generated geometries not only align visually with the input images but also accurately reflect their underlying scale, shape, and depth. 
During the conditioning training, we simulate real-world partial scans or estimated depth maps by rendering each 3D asset from multiple viewpoints, thereby obtaining corresponding RGB images, camera parameters, and ground-truth depth maps. These RGB images are then processed using advanced depth estimation techniques, including MoGe~\cite{wang2024moge} and Metric3D~\cite{yang2024depth}, to produce an estimated depth map and then projected as partial point clouds. To ensure scale consistency, we align estimated depth maps from MoGe and Metric3D with ground-truth depth maps by scaling and shifting them based on the median and median absolute deviation of valid depth values. The resulting point clouds are then normalized to a canonical [-1, 1]$^3$ space to ensure consistent spatial representation for coarse object alignment.


To bolster the model's robustness and its ability to generalize across diverse real-world scenarios, we employ a data augmentation strategy that interpolates between ground-truth partial point clouds $\bm{p}_\text{gt}$ (projected from ground truth depth map to simulate accurate depth) and noisier, estimated partial point clouds $\bm{p}_\text{est}$ (projected from estimated depth map and aligned to simulate estimated noisy depth from RGB). This interpolation is mathematically represented as:
 $  \bm{p}_\text{disturb} = \alpha\cdot \bm{p}_\text{gt} +(1-\alpha)\cdot \bm{p}_\text{est}$,
where $\alpha\in[0,1]$ is a weighting factor which is sampled uniformly during training. Our object generator, named ``\textit{ObjectGen}'', with partial point cloud conditioning is formulated as
\begin{equation}
    \epsilon (\bm{Z}_t; t,\bm{c},\bm{p}_\text{disturb}) \rightarrow \bm{Z},
\end{equation}
where the conditioning adaptation scheme is based on attention mechanism similar to \citet{zhang20233dshape2vecset,zhang2024clay}.
Additionally, to mimic real-world occlusions and missing data, we randomly mask sets of basic primitives—such as circles and rectangles—in the depth maps from various camera views. This results in partial point clouds with occluded and incomplete regions, further enhancing the model's ability to handle imperfect inputs.
A critical design choice in our approach is to maintain the alignment of partial point clouds with the geometry in our training data set. Unlike methods that apply random scaling, translation, or rotation to augmented point clouds, our aligned partial point clouds ensure that the generative model can more effectively conform to the input point clouds's inherent structure. This alignment restricts the model to adhere closely to the actual shapes and scales of objects, thereby facilitating more precise and coherent 3D reconstructions. By conditioning on these well-aligned partial point clouds, our model achieves superior alignment both in overall size and local geometric details, resulting in high-quality and reliable 3D geometry generation.

\subsection{Generative Alignment}
\label{sec:transformationgen}

Each generated 3D object is within a normalized volume and assumes a canonical pose that may not be aligned with the image and scene space point cloud. 
This is because the image conditions use high-level features, such as DINOv2, to achieve better generalization.
Ensuring that each object is correctly transformed and scaled to align with its presentation in the scene is crucial for scene composition.
Though traditional alignment methods, such as Iterative Closest Point (ICP)~\cite{arun1987least,best1992method}, can be employed, they often fail to account for semantic context, leading to frequent misalignments and diminished accuracy (see Fig.~\ref{Ablation_icp_dr}). 
Instead, we introduce an alignment generative model conditioned on the scene-space partial point cloud $\bm{q} \in \mathbb{R}^{N\times 3}$ and the canonical-space geometry latent code $\bm{Z}$. Formally, we define our alignment generator ``\textit{AlignGen}'' as:
\begin{equation}
    \epsilon_\text{align} (\bm{p}_t; t, \bm{q}, \bm{Z}) \rightarrow \bm{p},
\end{equation}
where $\epsilon_\text{align}$ is a point cloud diffusion transformer, $\bm{p} \in \mathbb{R}^{N\times 3}$ is the transformed version of the scene-space partial point cloud to the canonical space, aligning with the generated object mesh. $\bm{Z}$ is the generated geometry latent of object corresponding to $\bm{p}$ from the object generation model. 
$\bm{p}_t$ is the noised version of $\bm{p}$ at timestep $t$.
%
In essence, the generation model maps the scene-space partial point cloud $\bm{q}$ to $\bm{p}$ in the canonical $[-1,1]^3$ space, aligning it with the generated object mesh. We can subsequently recover the similarity transformation (i.e., scaling, rotation, and translation) from $\bm{q}$ and $\bm{p}$ using the Umeyama algorithm~\cite{umeyama1991least} as they are point-wise corresponded. This final step is numerically more stable than directly predicting transformation parameters.

In practice, we employ distinct conditioning strategies for the input point cloud $\bm{q}$ and the geometric latent $\bm{Z}$. For $\bm{q}$, we concatenate the input point cloud with the diffusion sample $\bm{p}_t$ along the feature channel dimension, enabling the transformer architecture to learn explicit correspondences between the noisy canonical-frame partial cloud and the world-space partial cloud.
For the geometric latent $\bm{Z}$, we apply a cross-attention mechanism to inject it into the point diffusion transformer. This approach ensures that the model effectively incorporates spatial and geometric relations.
Additionally, due to symmetry and replicated geometrical shapes, multiple valid $\bm{p}$ may exist for a given $\bm{q}$ and $\bm{Z}$. Our diffusion model addresses this by sampling multiple noise realizations and aggregating the resulting transformations, to select the most confident and coherent representations.

\subsection{Iterative Generation Procedure}

Recall that in our design, the object point cloud is unusable for object generation initially, as it is represented in the scene space, while our object generation model requires canonical-space point cloud for conditioning.
%
Solely depending on image cues for object generation often fails to produce pixel-aligned geometry, mainly because of the high-level semantic conditioning and biases inherent in 3D datasets.
Fortunately, our design enables seamless integration of the object generation and alignment modules through a joint, iterative process. 
This integration ensures that each generated 3D object is not only visually consistent with the input image but also accurately positioned and scaled within the scene. The iterative workflow with step index $k$ can be summarized in three key steps:

\noindent\paragraph{Step 1: Object Generation}
For an object image with mask, the \textbf{Object Generation} module (Sec.~\ref{sec:objectgen}) synthesizes the geometry latent code \(\bm{z}^{(k)}\) based on the image features \(\bm{c}\) derived from DINOv2 and the aligned point cloud \(\bm{p}^{(k)}\) in canonical coordinates. We set $\bm{p}^{(0)}$ to scene space point cloud $\bm{q}$ and set the point cloud conditioning scale factor \(\beta^{(k)}\) to progressively increase from 0 to 1 as iteration procedure goes on, allowing the partial point cloud to take influence over time. Formally, this process is represented as:
\begin{equation}
    \bm{z}^{(k)} = \text{ObjectGen}(\bm{c}, \bm{p}^{(k)} \otimes \beta^{(k)}).
\end{equation}
Hence our object generator solely relies on masked image conditioning in the first step. The latent code \(\bm{z}^{(k)}\) is then decoded into a 3D geometry using the VAE decoder \(\mathcal{D}\).

\noindent \paragraph{Step 2: Alignment}
Subsequently, the \textbf{Generative Alignment} module (Sec.~\ref{sec:transformationgen}) takes the newly generated geometry latent code \(\bm{z}^{(k)}\) and the partial point cloud \(\bm{q}\) in scene coordinates to predict a transformed canonical-space partial point cloud \(\bm{p}^{(k+1)}\):

\begin{equation}
\bm{p}^{(k+1)} = \text{AlignGen}(\bm{q}, \bm{z}^{(k)}).
\end{equation}

This transformed point cloud \(\bm{p}^{(k+1)}\) serves as an improved alignment reference for the next iteration. By leveraging the generative transformation model, the model ensures that the scaling, rotation, and translation adjustments are both precise and semantically informed.

\noindent \paragraph{Step 3: Refinement}
With the updated partial point cloud \(\bm{p}^{(k+1)}\), the system could estimate a new similarity transformation to refine the alignment of the generated geometry within the scene. This updated partial point cloud is then fed back into the \textbf{Object Generation} module for the next iteration, allowing for progressive enhancements in both geometry accuracy and spatial positioning.

This iterative loop—alternating between geometry generation and transformation estimation—continues until convergence criteria are met. Convergence is achieved when the changes in transformation parameters fall below a predefined threshold or when a maximum number of iterations is reached. The result is a high-fidelity 3D object that is both visually accurate and geometrically aligned with the input data.
%
By tightly integrating the \textbf{Object Generation} and \textbf{Alignment Generation} modules within an iterative framework, our approach effectively balances aesthetic fidelity with geometric precision. This joint generation process leverages both visual and depth information, ensuring that each 3D asset is of high quality and accurately positioned. Consequently, the pipeline lays a robust foundation for constructing physically correct and visually coherent 3D scenes, facilitating a wide range of downstream applications such as editing, rendering, and animation.

Once the object geometry is determined, we apply a state-of-the-art texture generation module to create photo-realistic surface details. Following established texture synthesis pipelines~\cite{zhang20233dshape2vecset,zhang2024clay}, we assign UV mappings and train a generative network to paint detailed textures onto the 3D meshes. This module is designed to robustly handle images under various augmentation, ensuring that the final textures match the input appearance even under occlusion or limited visibility.

%% file: sec/5_method_part3_jigu.tex
\begin{figure*} []
  \centering
  \includegraphics[width=\textwidth]{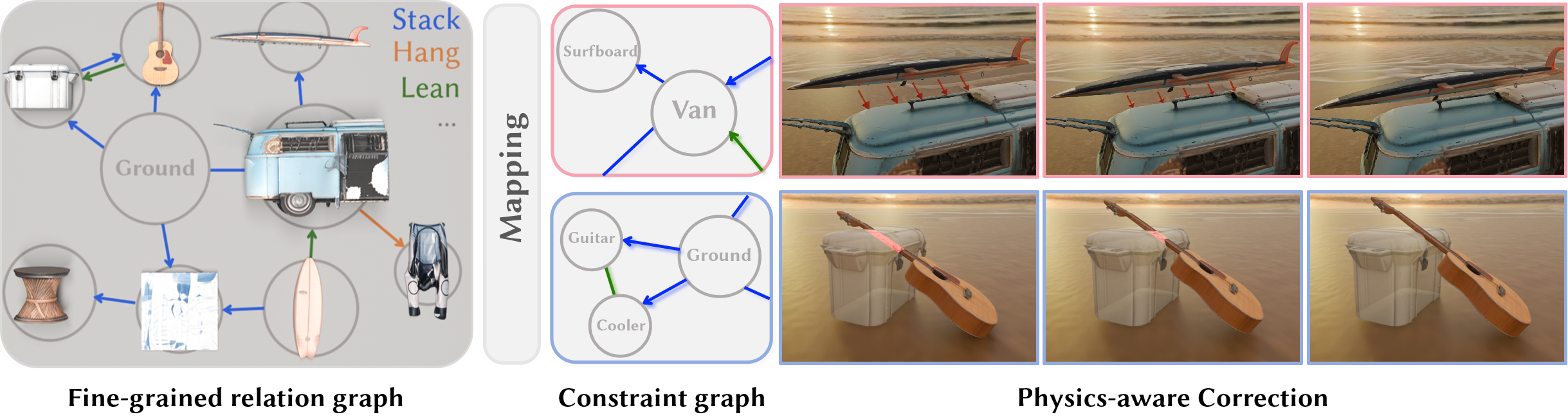}
  \captionof{figure}{Physics-aware correction via constraint graph mapped from fine-grained relation graph. Top: Floating surfboard grounded on the van. Bottom: Penetrating guitar and cooler separated.} \label{fig_simu_demonstrate}
\end{figure*}

\begin{figure*} []
  \centering
  \includegraphics[width=\textwidth]{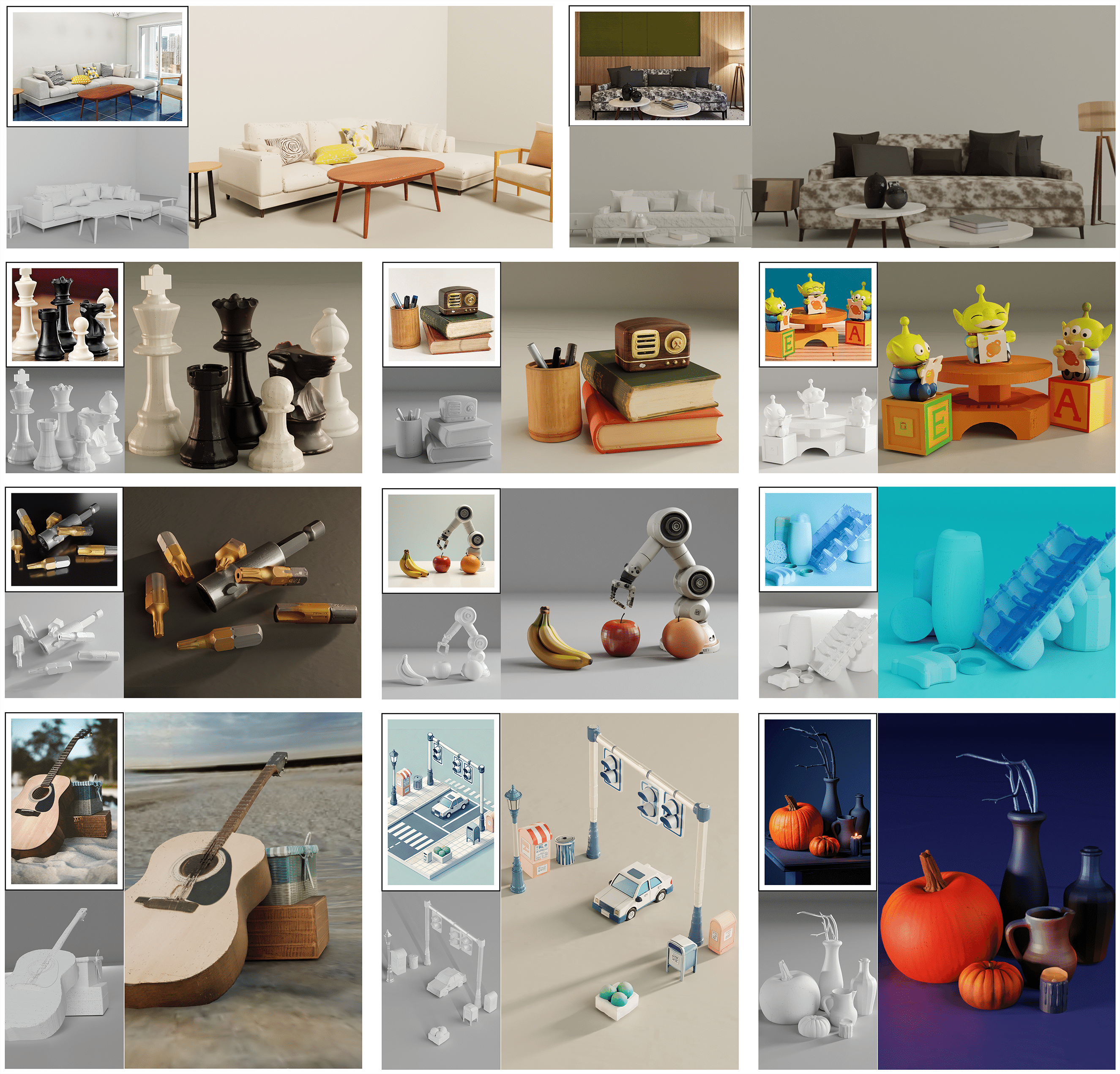}
  \captionof{figure}{Bringing the vibrant diversity of the real world into the virtual realm, this collection reimagines open-vocabulary scenes as immersive digital environments, capturing the richness and depth of each unique setting. For each scene, the images display as follows: the top-left shows the input image, the top-center displays the rendered geometry, and the right presents the rendered image with realistic textures.}
  \label{fig_gallery}
\end{figure*}

\section{Physics-Aware Correction}
\label{sec:physics-aware-correction}

The pipeline detailed in Sec.~\ref{sec:method2} individually generates each 3D object instance and estimates its similarity transformation (scaling, rotation, and translation) based on a single input image. While our proposed modules achieve high accuracy, the resulting scenes are sometimes not physically plausible. For instance, as illustrated in Fig.~\ref{fig_simu_demonstrate}, one object (e.g., a guitar) may intersect with another (e.g., a cooler), or an object (e.g., a surfboard) may appear to float unnaturally without any support (e.g., from a van). 

To address these issues, we introduce a physics-aware correction process that optimizes the rotation and translation of objects, ensuring the scene adheres to physical constraints consistent with common sense.
The correction process is motivated by physical simulation (Sec.~\ref{sec:rigid-body-sim}) and formulated as an optimization problem (Sec.~\ref{sec:physical-constraints}) based on inter-object relations represented by a scene graph (Sec.~\ref{sec:scene-relation-graph}) extracted from the image.

\subsection{A Quick Primer to Rigid-Body Simulation}
\label{sec:rigid-body-sim}

We introduce the fundamental principles of physical (rigid-body) simulation, which inspire our problem formulation and make our framework more accessible to downstream applications such as gaming and robotics. For a thorough survey, we refer the readers to \citet{BenderETC12}.

In rigid-body simulations, the world is modeled as an ordinary differential equation (ODE) process. In each simulation step, it begins with the Newton-Euler (differential) equations, which describe the dynamic motion of rigid bodies in the absence of contact. \emph{Collision detection} is conducted to find the contact points between rigid bodies, which are needed to determine contact forces. For contact handling and collision resolution, there are usually several conditions: non-penetration constraints to prevent bodies from overlapping, a friction model ensuring contact forces remain within their friction cones, and complementarity constraints that enforce specific disjunctive relations among variables. Solvers are used to resolve the system comprising equations and inequalities, subsequently updating the velocity and position of each rigid body.

A straightforward approach to enhance physical plausibility is to utilize an off-the-shelf rigid-body simulator to process the scene, starting from the initial state estimated by the pipeline previously described and obtaining the rest state after simulation. However, this method presents several challenges.
\\ 1) \emph{Partial Scene}: Some objects may be missing due to the limitations of 2D foundation models, and thus not reconstructed. Simulating a partial scene under full physical rules can lead to suboptimal results (see Fig.~\ref{Ablation_simu}). 
\\2) \emph{Imperfect Geometries}: While our 3D generative model produces high-quality geometries, minor imperfections may still occur. Rigid-body simulators typically require convex decomposition~\cite{mamou2009simple,mamou2016volumetric,wei2022approximate} of objects, which introduces additional complexity and hyperparameters. Overly fine-grained decomposition can result in non-flat, complex surfaces, causing objects to fall or move unexpectedly during simulation. Conversely, coarse decomposition may lead to visually floating objects due to discrepancies between the visual and collision geometries.
\\3) \emph{Initial Penetrations}: Despite the high accuracy of pose estimation, significant inter-object penetrations may exist in the initial state. These penetrations create instability for standard rigid-body solvers and, in some cases, lead to unsolvable scenarios if the solver is not customized for those cases.

Thus, we propose a customized and simplified ``physical simulation'' to optimize the object poses, ensuring that the scene adheres to common-sense physical principles derived from the single image.
\emph{Note that our approach does not model full dynamics. For example, an object may not remain stable in its current pose over time. However, it should be physically plausible at the current time step. We argue that our optimized results can serve as a reliable initialization for subsequent physical simulations.}

\subsection{Problem Formulation and Physical Constraints}
\label{sec:physical-constraints}

We formulate the physics-aware correction process as an optimization problem, aiming to minimize the total cost that represents pairwise constraints on objects.

\begin{equation}
\label{eq:physics-aware-objective}
    \min_{\mathcal{T}=\{T_1,T_2,\dots,T_N\}} \sum_{i,j}C(T_i,T_j;\bm{o}_i,\bm{o}_j)
\end{equation}
where $N$ is the number of objects, $T_i$ is the rigid transformation (rotation and translation) of the i-th object $\bm{o}_i$. $C$ is the cost function representing the relation between $\bm{o}_i$ and $\bm{o}_j$. Note that the cost function varies depending on the type of relation.

Motivated by physical simulation, we categorize the relations into two types: \emph{contact} and \emph{support}. The relations are identified with the assistance of a VLM, as detailed in Sec.~\ref{sec:scene-relation-graph}.

1) \emph{Contact} describes whether two objects $\bm{o}_i$ and $\bm{o}_j$ are in contact. Let $D_i(p)$ denote the signed distance function induced by $\bm{o}_i$ at the point $p$, which is used to define the constraint. $D_i(p)=D_j(p)=0$ indicates that $p$ is a contact point of $\bm{o}_i$ and $\bm{o}_j$. When $D_i(p)=0$ ($p$ is a surface point of $\bm{o}_i$), $D_j(p)<0$ indicates inter-object penetration, while $D_j(p)>0$ means the objects are separated. Thus, the cost function can be defined as:

\begin{equation}
\label{eq:contact-constraint}
\begin{split}
    C(T_i, T_j; \bm{o}_i \to \bm{o}_j) = 
    -\frac{\sum_{p \in \partial \bm{o}_j} D_i(p(T_j)) \mathbb{I}(D_i(p(T_j))<0)}{\sum_{p \in \partial \bm{o}_j} \mathbb{I}(D_i(p(T_j))<0)} 
    \\
    + \max(\min_{p \in \partial \bm{o}_j}D_i(p(T_j)),0)
    \\
    C(T_i, T_j; \bm{o}_j \to \bm{o}_i) =-\frac{\sum_{p \in \partial \bm{o}_i} D_j(p(T_i)) \mathbb{I}(D_j(p(T_i))<0)}{\sum_{p \in \partial \bm{o}_i} \mathbb{I}(D_j(p(T_i))<0)}
    \\
    + \max(\min_{p \in \partial \bm{o}_i}D_j(p(T_i)),0)
    \\
    C(T_i,T_j) = C(T_i, T_j; \bm{o}_i \to \bm{o}_j) + C(T_i, T_j; \bm{o}_j \to \bm{o}_i)
    \\
    \text{if } \bm{o}_i \text{ and } \bm{o}_j \text{ are in contact}
\end{split}
\end{equation}
where $\partial \bm{o}_i$ denotes the surface of $\bm{o}_i$, and $\mathbb{I}$ is the indicator function. The constraint ensures that there is no penetration and at least one contact point between the objects. Note that $p \in \partial \bm{o}_i$ is a function of $T_i$. The contact constraint defined here is bilateral, meaning it applies to both objects.

2) \emph{Support} is a unilateral constraint, which is a special case of \emph{Contact}. If $\bm{o}_i$ supports $\bm{o}_j$, it implies that the pose $T_j$ of $\bm{o}_j$ should be optimized while $\bm{o}_i$ is assumed to be static. This scenario typically occurs when multiple objects are stacked vertically. The cost function for this case is similar to the one in \emph{Contact}, but it only involves one direction:

\begin{equation}
\label{eq:support-constraint}
    C(T_i,T_j) = |\min_{p \in \partial \bm{o}_j} D_i(p(T_j))|,\ \text{if } \bm{o}_i \text{ supports } \bm{o}_j
\end{equation}


Furthermore, for flat supporting surfaces like the ground or walls, we regularize the SDF values near the contact region, to ensure that objects make close contact with these surfaces. This regularization handles scenarios where objects are partially reconstructed, such as a van with only two wheels, as illustrated in Fig.~\ref{fig_simu_demonstrate}.
\begin{equation}
\label{eq:support-constraint3}
    C(T_i,T_j) = \frac{\sum_{p \in \partial \bm{o}_j} D_i(p(T_j) \mathbb{I}(0<D_i(p)<\sigma)}{\sum_{p \in \partial \bm{o}_j} \mathbb{I}(0<D_i(p)<\sigma)}
\end{equation}
where $\mathbb{I}$ is the indicator function, and $\sigma$ is a threshold to decide whether a point is sufficiently close to the surface.



\subsection{Scene Relation Graph}
\label{sec:scene-relation-graph}

Physical cues, particularly inter-object relations, are visually present in the image.  We leverage the strong common-sense reasoning capabilities~\cite{rana2023sayplan,li2024llm,cheng2024navila} of visual-language models, specifically GPT-4v~\cite{achiam2023gpt}, to identify pairwise physical constraints as defined in Sec.~\ref{sec:physical-constraints}. Given an image, we employ the Set of Mark~\cite{yang2023setofmark} (SoM) technique to visually prompt GPT-4v to describe the inter-object relations, and subsequently extract a \emph{scene relation graph} from the answers. To address the sampling uncertainty inherent in VLMs, we adopt an ensemble strategy, combining results from multiple trials. We define relations as correct if they appear in more than half of the samples to produce a robust inferred graph. To be more specific, we apply the Set-of-Mark method with random colorization and numerical ordering multiple times,enabling more reliable and consistent outputs for further GPT-based question-answering tasks.
 %

Instead of directly asking GPT-4v to identify \emph{Support} and \emph{Contact} relations, we first provide it with more fine-grained physical relations, such as \emph{Stack} (Object 2 supports Object 1), \emph{Lean} (Object 1 leans against Object 2), and \emph{Hang} (Object 2 supports Object 1 from above). We instruct GPT-4v to analyze numbered objects from the Set-of-Mark method and output all contact-based relations, covering six types: Stack, Lean, Hang, Clamped, Contained, and Edge/Point. The prompt specifies that only contacting objects have relations and defaults to Stack for ambiguous cases.

We then map these detailed relations to the predefined categories of \emph{Support} and \emph{Contact} for further optimization. Specifically, if there are edges pointing toward each other between two nodes, the edge is categorized as \emph{Contact}; otherwise, it is categorized as \emph{Support}. Prompting GPT-4v with these nuanced relations helps eliminate potential ambiguity in binary relation classification and facilitates more accurate reasoning by GPT-4v.
An example of the resulting graph is illustrated in Fig.~\ref{fig_simu_demonstrate}.

The mapped scene constraint graph is a directed graph where nodes represent object instances and edges denote physical relations between objects. A \emph{Contact} relation is represented by a bidirectional edge, while a \emph{Support} relation is depicted as a directed edge. This graph serves as the foundation for defining the cost functions used in Eq.~\ref{eq:physics-aware-objective}.

\subsection{Optimization with Physics-Aware Relation Graph}

Given the physical constraints defined by the inferred relation graph, we can instantiate our cost functions as described in Eq.~\ref{eq:physics-aware-objective}. The graph allows us to reduce the number of pairwise constraints that need to be optimized, in contrast to a full physical simulation.

For the implementation, we uniformly sample a fixed number of points from the surface of each object at its rest pose. These points are then transformed according to the current object’s pose parameters and used to query the SDF values with respect to another object (and its pose). SDF computation is handled by Open3D, and Pytorch is used to auto-differentiate the loss function.

%% file: sec/6_exp.tex
\begin{figure*} []
  \centering
  \includegraphics[width=\textwidth]{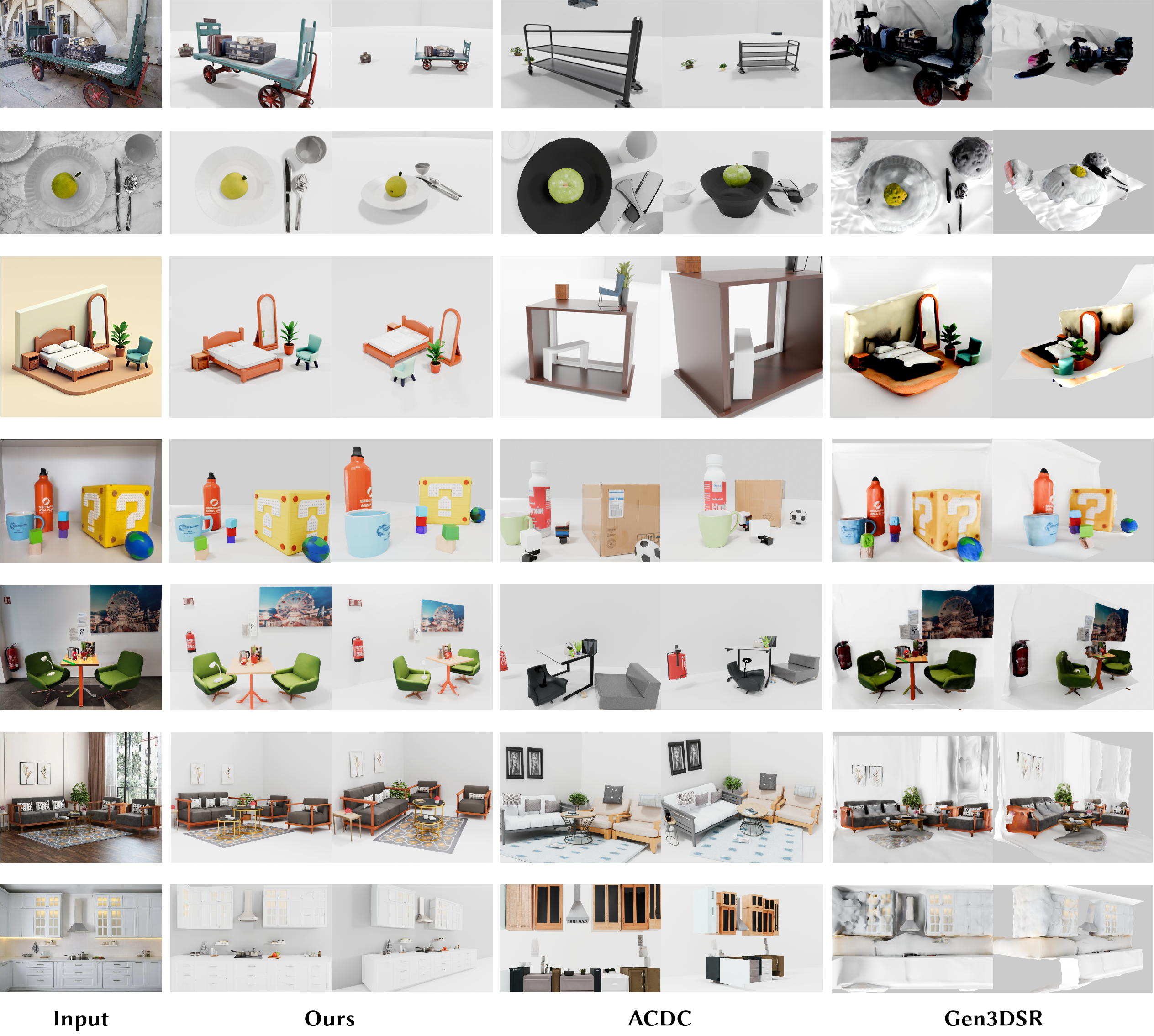}
  \captionof{figure}{Qualitative comparisons of CAST with state-of-the-art single-image scene reconstruction methods. From left to right: Input image, CAST, ACDC, and Gen3DSR. Top to bottom: random open vocabulary dataset (rows 1–3), Gen3DSR input (rows 4–5), ACDC input (rows 6–7).}
  \label{fig_comparison1}
\end{figure*}

\section{Result}
\label{sec:experiments}

Fig.~\ref{fig_gallery} showcases a range of 3D scenes generated by our method from single-view inputs across a diverse range of open-vocabulary scenarios, featuring detailed indoor environments, close-up captures of objects, and AI-generated imagery. These examples highlight the versatility and robustness of our approach, exhibiting high-fidelity geometry, realistic textures, and convincing scene compositions.

\subsection{Implementation Details}
\paragraph{ObjectGen}
The ObjectGen (Sec.~\ref{sec:objectgen}) model's pretraining follows the methodology outlined in 3DShape2VecSet~\cite{zhang20233dshape2vecset} and CLAY~\cite{zhang2024clay}, where we leverage both a Variational Autoencoder (VAE) and a Latent Diffusion Model (LDM) to generate 3D object geometries. Both the VAE and LDM modules are implemented using a 24-layer transformer, comprising a total of 1.5 billion parameters. The model is trained on the Objaverse~\cite{deitke2023objaverse} dataset, which consists of approximately 500,000 3D assets after filtering. 
The partial point cloud conditioning follows a similar approach to CLAY's adaptation framework. We encode the canonical-space partial point cloud as positional embeddings with a feature dimension of 512, which are injected into the main LDM transformer using cross-attention mechanisms. For each 3D asset, we render 32 views and precompute depth maps using MoGe~\cite{wang2024moge} and Metric3D~\cite{yin2023metric3d}. These depth maps are then lifted into point clouds during training, with random masks applied to simulate occlusions. We sample 2048 points from the unprojected point cloud using Farthest Point Sampling (FPS), which serves as conditioning input for the LDM. To enhance the model's robustness, we randomly interpolate between the ground truth and predicted partial point clouds, allowing the system to handle data of varying quality. The conditioning module is trained on 200K curated data from Objaverse over 3000 epochs with 64 Nvidia A800 GPUs required approximately one week. The AdamW optimizer is used with a learning rate of 1e-5. For inferencing a single object, object generation takes approximately 7 seconds, and texture generation takes approximately 10 seconds per object on an NVIDIA A6000 GPU.

\paragraph{AlignGen}
The AlignGen (Sec.~\ref{sec:transformationgen}) module, responsible for generating pose alignment, utilizes a 24-layer transformer with a feature dimension of 512, resulting in a total of 150 million parameters. 
During training, we randomly sample a partial point cloud in the canonical space from the point clouds lifted from precomputed depth maps and apply a random transformation to this point cloud. The transformed point cloud, along with the geometry latent code $\bm{z}$ from ObjectGen, is used as the conditioning input.
FPS is utilized to sample 2048 points from the partial point cloud to ensure a fixed number of inputs to the transformer.
Training is conducted on the same 200K curated dataset over 1,500 epochs with 64 Nvidia A800 GPUs for approximately two day. The AdamW optimizer is used with a learning rate of 1e-5. During inference, AlignGen module takes around 1 second for one object for pose generation.

\begin{table}[]
\centering
\caption{Quantitative comparison of scene reconstruction methods across four metrics with CLIP score, GPT-4 ranking, user study of visual quality (VQ), and physical plausibility (PP).}
\label{tab:performance_comparison_visual}
\begin{tabular}{lcccc}
\toprule
\textbf{Method}& CLIP↑ & GPT-4↓ &  VQ↑  & PP↑   \\ \midrule
ACDC     & 69.77  & 2.7   &  5.58\% &  22.86\%  \\
Gen3DSR  & 79.84  & 2.175   &  6.35\%  & 5.72\% \\
ours     & \textbf{85.77}  & \textbf{1.125}   &  \textbf{88.07\%} & \textbf{71.42\%} \\ \bottomrule
\end{tabular}
\end{table}

\subsection{Comparison}

\begin{table}[]
\centering
\caption{Quantitative comparison of scene reconstruction performance on the 3D-Front indoor dataset. We evaluate different methods based on Chamfer Distance (CD) for shape accuracy, F-Score (FS) for object-level reconstruction quality, and Intersection over Union (IoU) for scene-level overlap. }
\label{tab:performance_comparison_3Dfront}
\begin{tabular}{lccccc}
\toprule
\textbf{Method} & CD-S↓ & FS-S↑ & CD-O↓ & FS-O↑ & IoU-B↑ \\
\midrule
ACDC  & 0.104 & 39.46 & 0.072 & 41.99 & 0.541\\
InstPIFU  & 0.092 & 39.12 & 0.103 & 38.29 & 0.436\\
Gen3DSR  & 0.083 & 38.95 & 0.071 & 39.13 & 0.459 \\
ours      & \textbf{0.052} & \textbf{56.18} & \textbf{0.057} & \textbf{56.50} & \textbf{0.603} \\
\bottomrule
\end{tabular}
\end{table}

\paragraph{Qualitative Comparisons}
We first evaluate our method, CAST, against state-of-the-art single-image scene reconstruction techniques on open-vocabulary scenarios. We also included images used by ACDC and Gen3DSR to further demonstrate the scene reconstruction results of different methods. Fig.~\ref{fig_comparison1} illustrates the performance of three methods—(1) the retrieval-based approach ACDC~\cite{dai2024automated}, (2) the generation-based method Gen3DSR~\cite{dogaru2024generalizable}, and (3) our proposed CAST—across both reference and novel views. Our results highlight CAST’s superior ability to accurately reconstruct scenes in diverse settings, including indoor and outdoor environments, close-up perspectives, and AI-generated imagery.
%

As shown in Fig.\ref{fig_comparison1}, CAST distinguishes itself from both ACDC and Gen3DSR through innovative advancements. Unlike ACDC, which is limited to indoor scenes and relies on large datasets for object retrieval, often producing objects similar to those in the scene rather than the objects themselves, CAST supports open-vocabulary generalization. This allows CAST to accurately reconstruct objects in varied and complex environments. While ACDC uses simple bounding boxes as proxies, CAST combines image-based physical priors with mesh optimization to effectively manage complex scenes.
In contrast to Gen3DSR, CAST employs direct 3D generation via a Masked Autoencoder, eliminating the error-prone 2D inpainting step. This results in smoother meshes, significantly outperforming Gen3DSR in single-object generation quality, particularly in challenging scenes. Moreover, Gen3DSR’s lack of simulation often leads to issues such as interpenetration or floating objects, causing scenes to appear consistent only from the input viewpoint and degrading novel view rendering. CAST, by contrast, ensures robust scene consistency across perspectives. CAST demonstrates robust scene reconstructions under varying conditions, underscoring its versatility for a broad range of real-world and generated scenarios.

To assess both the visual fidelity and semantic accuracy of the generated scenes, we employ two complementary evaluation methods including CLIP Score~\cite{taited2023CLIPScore} and GPT-4 Reasoning.
We compute the CLIP score between the rendered scene and the input image to measure overall reconstruction quality and visual similarity. To minimize environmental distractions, we remove backgrounds from both the rendered and reference images before computing the score.
We additionally leverage GPT-4 to rank the generated scenes based on various semantic aspects, including object arrangement, physical relations, and scene realism. This semantic feedback helps identify alignment or contextual errors that might not be apparent through pixel-based scores alone.

Beyond automated metrics, we also conducted a user study focusing on two key aspects including Visual Quality (VQ) and Physical Plausibility (PP).
We randomly selected paired reference, novel, and target views, asking participants to choose which method’s output best matched the input image in terms of both similarity and overall aesthetics.
To reduce potential biases introduced by visual resemblance, participants in a separate session only viewed rendered results—without the original input images—and judged which scene appeared more realistic based on physical constraints and common sense (e.g., preventing floating objects or improbable contacts).

As shown in Tab.~\ref{tab:performance_comparison_visual}, CAST outperforms both ACDC and Gen3DSR in all four evaluated metrics, confirming its effectiveness in producing scenes that are visually coherent and physically plausible.

\begin{figure} []
  \centering
  \includegraphics[width=\columnwidth]{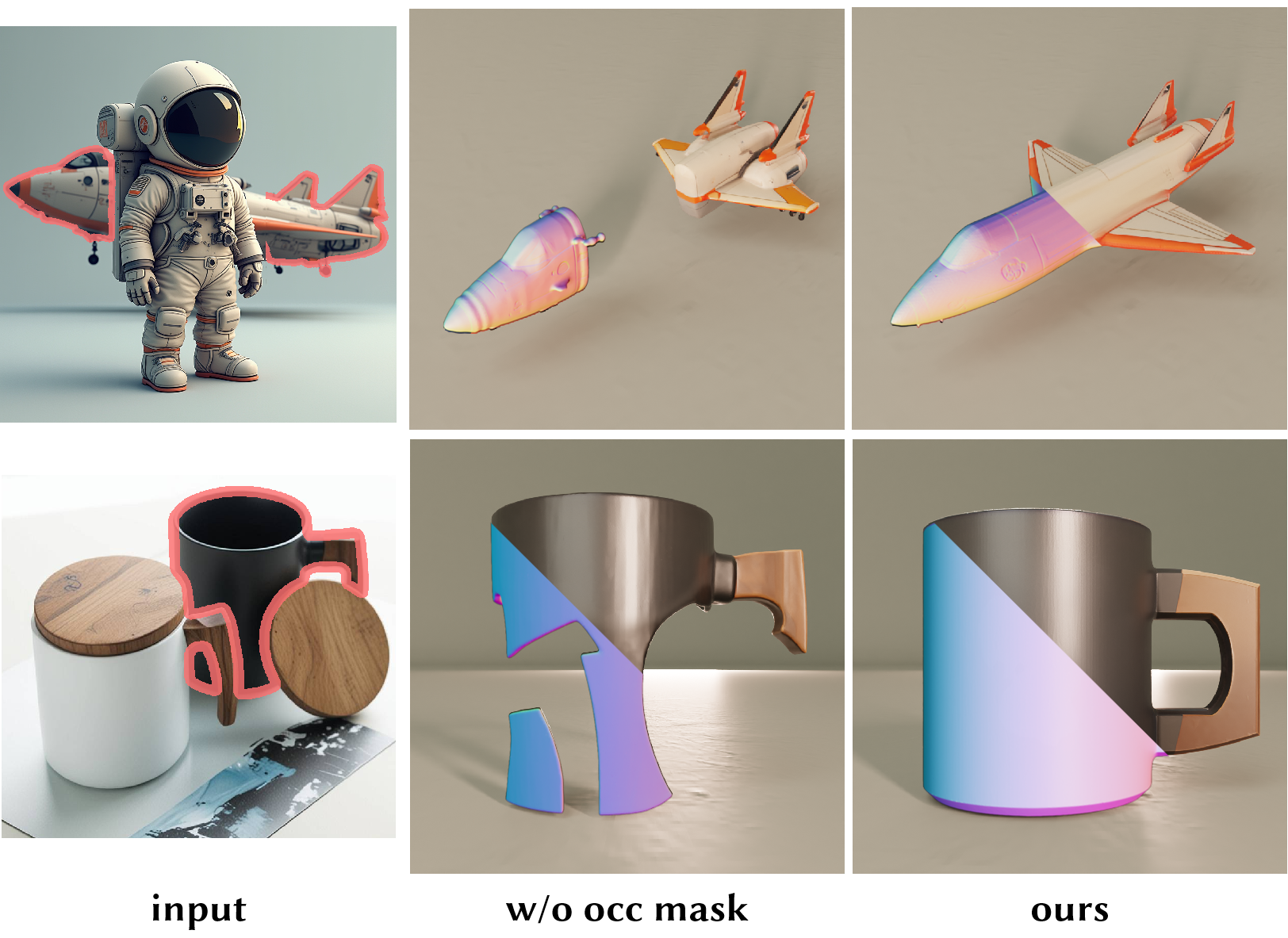}
  \captionof{figure}{We evaluate the generation performance with and without the occlusion-aware generation module. The RGB and normal renderings of the object highlight the significance of this module in ensuring the completeness and high quality of the generated object.} 
  \label{Ablation_occlusion}
\end{figure}

\paragraph{Quantitative Comparisons}

%
Although CAST is designed to handle open-vocabulary scenes, many such scenes lack mesh ground truth, which complicates direct quantitative comparisons. To address this, we perform additional evaluations on the 3DFront dataset~\cite{fu20213d}. This dataset offers ground-truth meshes alongside corresponding rendered images, enabling a more precise assessment of both object-level and scene-level reconstructions.
We compare our method with InstPIFu~\cite{liu2022towards}, ACDC~\cite{dai2024automated}, and Gen3DSR~\cite{dogaru2024generalizable}.
We compute Chamfer Distance and F-Score at the object level, as well as IoU, Chamfer Distance, and F-Score at the scene level, to assess both the fidelity of individual object geometries and the accuracy of their spatial layout.
%
To ensure fairness, we replace the segmentation modules in other methods with ground-truth (GT) masks so that any differences stem purely from reconstruction ability rather than object partitioning.

As summarized in Tab.~\ref{tab:performance_comparison_3Dfront}, CAST not only achieves higher object-level generation quality but also surpasses existing approaches in scene layout accuracy. Even within the constraints of an indoor dataset, our method demonstrates robust performance and consistent improvements over competing baselines.

\subsection{Evaluations}


\begin{figure} []
  \centering
  \includegraphics[width=\columnwidth]{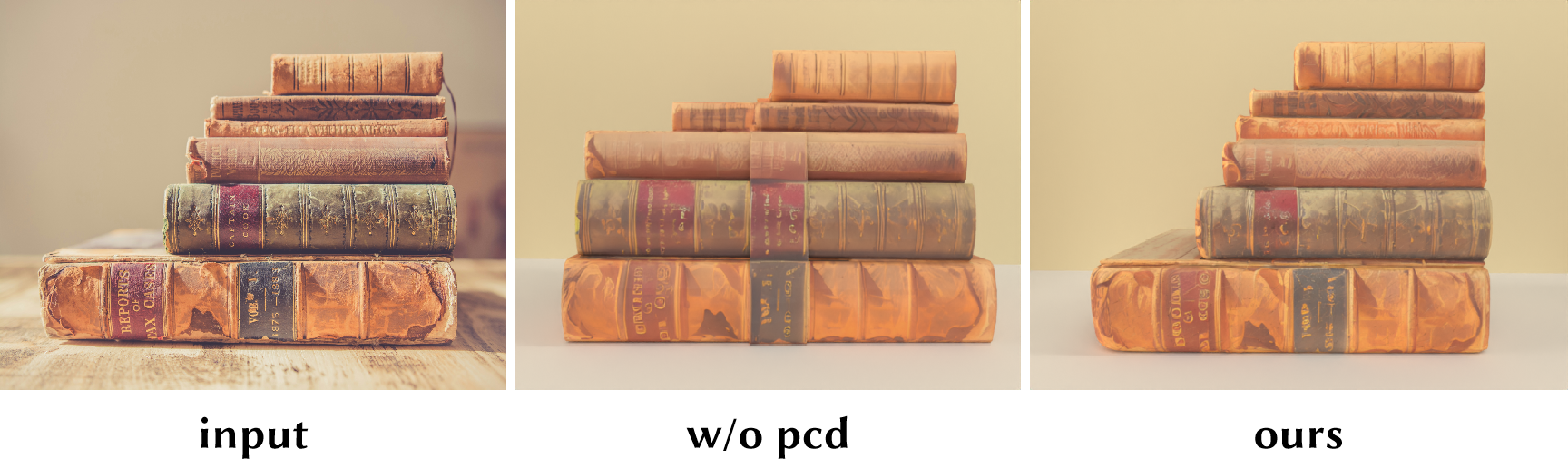}
  \captionof{figure}{A stack of books with varying lengths and widths directly generated as a single complex object, demonstrating how point cloud conditioning enhances the preservation of scale, dimensions, and local details compared to traditional methods.} 
  \label{Ablation_pointscondition}
\end{figure}

\begin{figure} []
  \centering
  \includegraphics[width=\columnwidth]{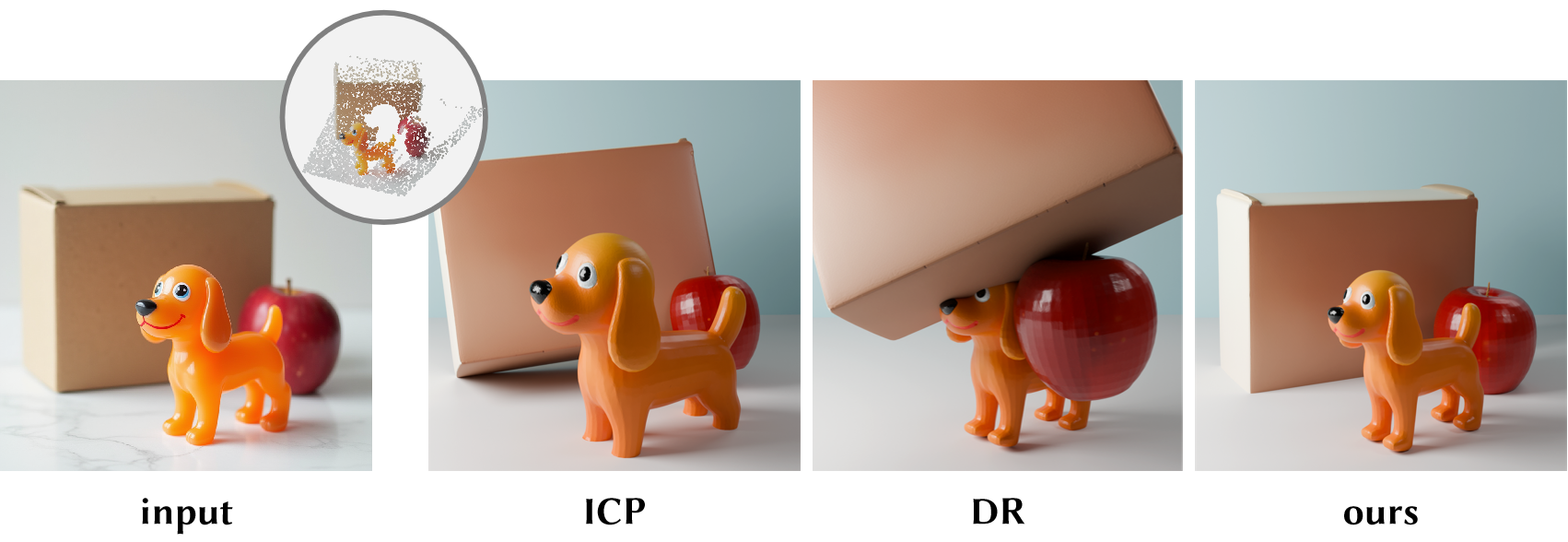}
  \captionof{figure}{Comparative evaluation of pose estimation methods. Our pose alignment module demonstrates superior alignment accuracy compared to Iterative Closest Point (ICP) and differentiable rendering (DR).} 
  \label{Ablation_icp_dr}
\end{figure}

To elucidate the individual contributions of key components in \methodname, we conducted a series of ablation studies. These experiments systematically removed or altered specific components to assess their impact on overall performance. The ablation studies focus on several key design choices: occlusion-aware object generation, point-cloud conditioning, generative alignment, and the physics-aware correction process.

\paragraph{Ablation on Occlusion-Aware Generation}
Occlusions are a significant challenge in scenes with complex objects. To assess the effectiveness of the Masked Autoencoder (MAE) in handling occlusions, we performed an ablation study comparing generation results with and without MAE components. As shown in Fig.~\ref{Ablation_occlusion}, the results highlight the importance of the occlusion-aware module. Without MAE, the generated objects for partially occluded regions exhibit significant degradation. For example, the spaceship appears fragmented and incomplete, while the cup is depicted as broken with missing parts. In contrast, when MAE conditioning is applied, the model successfully infers and fills the occluded regions, resulting in more accurate and visually coherent generations that align better with the input image. This demonstrates the critical role of the occlusion-aware module in ensuring that occluded objects are reconstructed accurately, improving both the completeness and realism of the final 3D scene.

\paragraph{Ablation on Partial Point Cloud Conditioning}
We conducted an ablation study to investigate the role of canonical space partial point cloud conditioning in our generation process. Although directly generating from the input image can produce visually plausible results. In the absence of pixel-level alignment, the model struggles to maintain correct object quantity and scale, resulting in unsatisfactory generation. To more effectively showcase the importance of point cloud conditioning in generating a single instance, we opted to directly generate a more complex instance structure: a stack of six books with varying lengths and widths. As depicted in Fig.~\ref{Ablation_pointscondition}. When the generation process relies solely on the input image, without the benefit of point cloud conditioning, the results frequently exhibit inaccuracies in both the number and dimensions of the generated objects. By contrast, integrating point cloud conditioning introduces a robust geometric prior that significantly improves the precision of the generated scene. This enhancement ensures that objects with intricate shapes and varying dimensions are reconstructed more accurately, closely resembling their real-world counterparts depicted in the input image. This demonstrates the critical role of geometric priors in enhancing the fidelity of 3D scene generation by preserving the true dimensions and shapes.


\begin{figure} []
  \centering
  \includegraphics[width=\columnwidth]{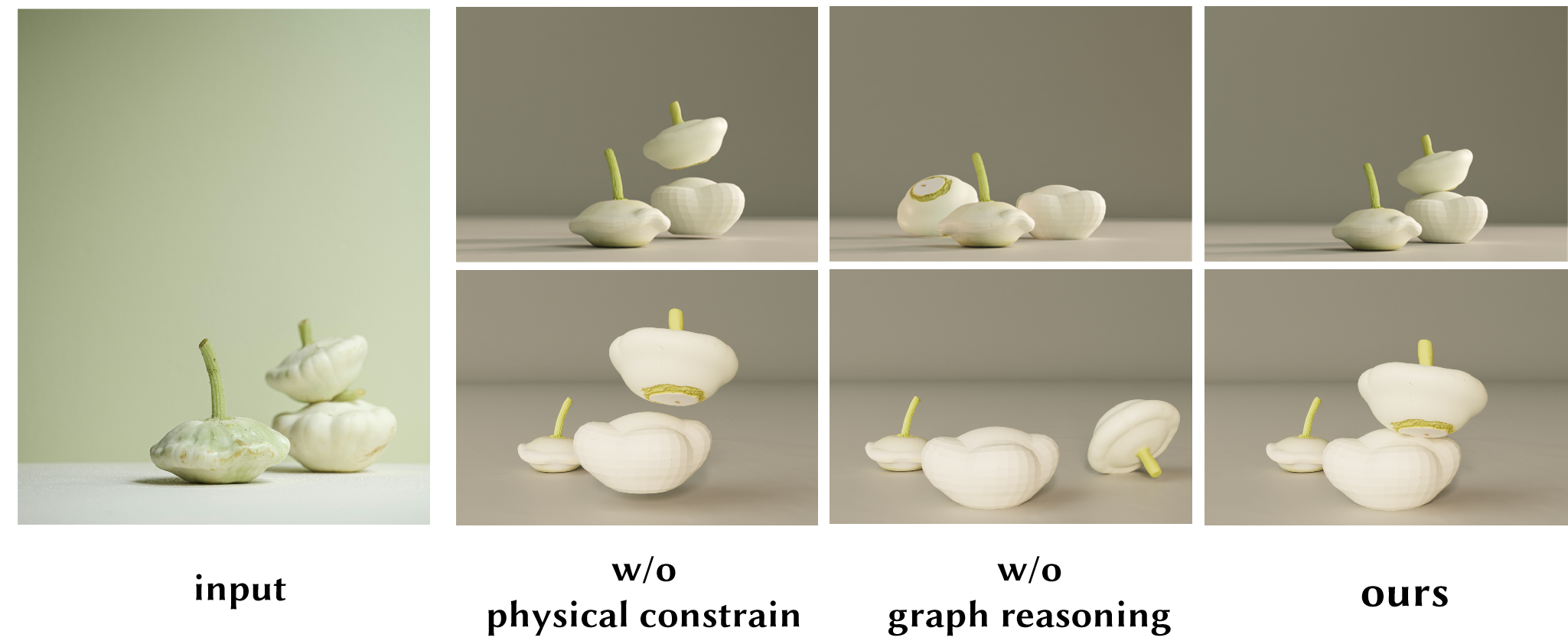}
  \captionof{figure}{Comparison of scene reconstruction with and without relational graph constraints. By integrating relational graph constraints, our method ensures both physical plausibility and accurate alignment with the intended scene, maintaining correct spatial relations.} 
  \label{Ablation_simu}
\end{figure}

\paragraph{Effectiveness of Alignment Generation}

To assess the effectiveness of our pose alignment module, we compared it with common pose estimation methods such as Iterative Closest Point (ICP)~\cite{arun1987least,best1992method} and differentiable rendering~\cite{nvdiffrast}. The generated mesh was provided to different pose estimation methods to align it with the reference RGB image and its corresponding depth prediction.
For the ICP method, we uniformly sampled a point cloud from the generated mesh and normalized both the sampled and estimated point clouds by their bounding boxes to address scale differences. We used the ICP implementation in Open3D~\cite{Zhou2018Open3DAM} to register these two normalized point clouds.
For differentiable rendering, we optimized the rotation and translation parameters to transform the generated mesh so that the rendered image aligned with the reference RGB image.
As shown in Fig.~\ref{Ablation_icp_dr}, our method surpasses both ICP and differentiable rendering in alignment accuracy. ICP often struggles with accurate pose estimation due to outliers in point clouds, unknown object scales, and symmetrical or repetitive geometries, which can lead to local minima. Differentiable rendering, on the other hand, is significantly impacted by occlusions in the RGB input, disrupting the optimization of object poses and preventing precise alignment with the input image. Our results show that our pose alignment module outperforms traditional ICP and differentiable rendering methods, demonstrating its robustness in accurately estimating object poses from generated meshes and improving alignment with input images.


\paragraph{Effect of Physical Consistency Enforcement}
In CAST, physical constraints are essential for achieving realistic object interactions and maintaining spatial coherence within a scene. While we address common challenges such as occlusions and incomplete views, issues like floating objects, penetration, and misaligned spatial relations still occur. As shown in Fig.~\ref{Ablation_simu}, scenes generated without relational constraints may appear physically inconsistent, when only physical simulation is applied, objects adhere to physical laws, but their relative positioning and overall arrangement can differ significantly from the intended scene (e.g., an onion might fall off a surface, disrupting the original composition).  By incorporating relational graph constraints, our method ensures that the objects not only comply with physical feasibility but also align with the intended scene layout, preserving both the physical plausibility and the desired spatial relations.


\begin{table}[]
\centering
\caption{Quantitative ablation study of the MAE module, point cloud conditioning (PCD), and the iterative refinement strategy (iter.). For simplicity, we only display the added key component in each row.}
\label{tab:ablation_occ_iter}
\begin{tabular}{lccccc}
\toprule
\textbf{Method} & CD-S↓ & FS-S↑ & CD-O↓ & FS-O↑ & IoU-B↑ \\
\midrule
Vanilla & 0.079 & 53.38 & 0.069 & 52.83 & 0.515\\
+ MAE  & 0.064 & 53.79& 0.066 &  54.32& 0.548\\
+ PCD   & 0.056 & 53.91 & 0.060 & 54.60  & 0.582 \\
+ iter.     & \textbf{0.052} &  \textbf{56.18}  & \textbf{0.057} & \textbf{56.50} & \textbf{0.603} \\

\bottomrule
\end{tabular}
\end{table}

\paragraph{Quantitative Ablation Study of Different Modules}
To quantitatively evaluate the contribution of each module to overall performance, we conducted a comprehensive ablation study. As shown in Tab.~\ref{tab:ablation_occ_iter}, we assessed the impact of removing or altering key components on the final scene quality. The results indicate that each component contributes significantly to the overall performance of our method. The quantitative analysis further highlights the importance of each module in achieving high-quality, physically consistent, and realistic scene reconstructions.

\begin{figure} []
  \centering
  \includegraphics[width=\columnwidth]{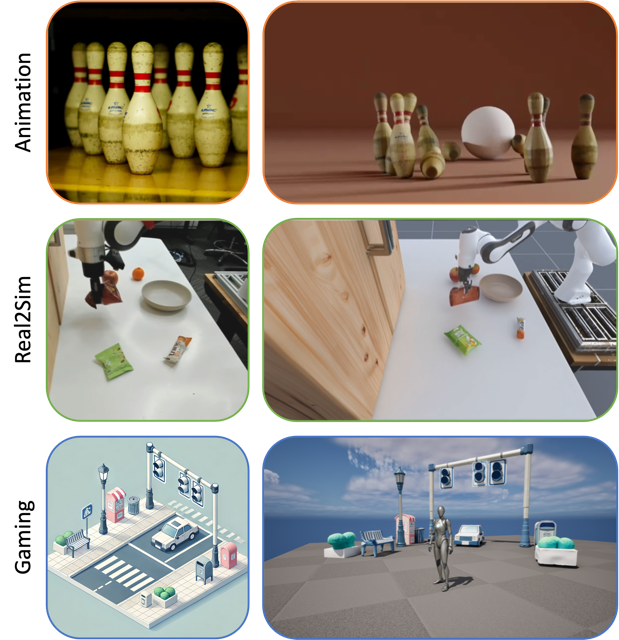}
  \captionof{figure}{CAST enables realistic physics-based animations, immersive game environments, and efficient real-to-simulation transitions, driving innovation across various fields.} 
  \label{fig_application}
\end{figure}

\paragraph{Applications}
As shown in Fig.~\ref{fig_application}, CAST transforms a single image into a fully realized 3D scene, enabling a wide range of applications. This ability to reconstruct detailed environments powers physics-based animation by ensuring realistic object interactions. It also supports real-to-simulation workflows in robotics, allowing for accurate scene replication from real-world datasets. In game development, CAST facilitates the creation of immersive environments, where faithfully reconstructed scenes are seamlessly integrated into interactive worlds using Unreal Engine.

%% file: sec/7_conclusion.tex
In this paper, we introduced CAST, a novel single-image 3D scene reconstruction method that combines geometric fidelity, pixel-level alignment, and physically grounded constraints. By integrating scene decomposition, a perceptive 3D instance generation framework, and physical correction techniques, CAST addresses key challenges such as pose misalignment, object interdependencies, and partial occlusions. This structured pipeline results in 3D scenes that are both visually accurate and physically consistent, pushing beyond the limitations of traditional object-centric approaches. We validated CAST through extensive experiments and user studies, demonstrating significant performance improvements over state-of-the-art methods in terms of visual quality and physical plausibility. We anticipate that CAST will serve as a strong foundation for future developments in 3D generation, scene reconstruction, and immersive content creation.

\begin{figure} []
  \centering
  \includegraphics[width=\columnwidth]{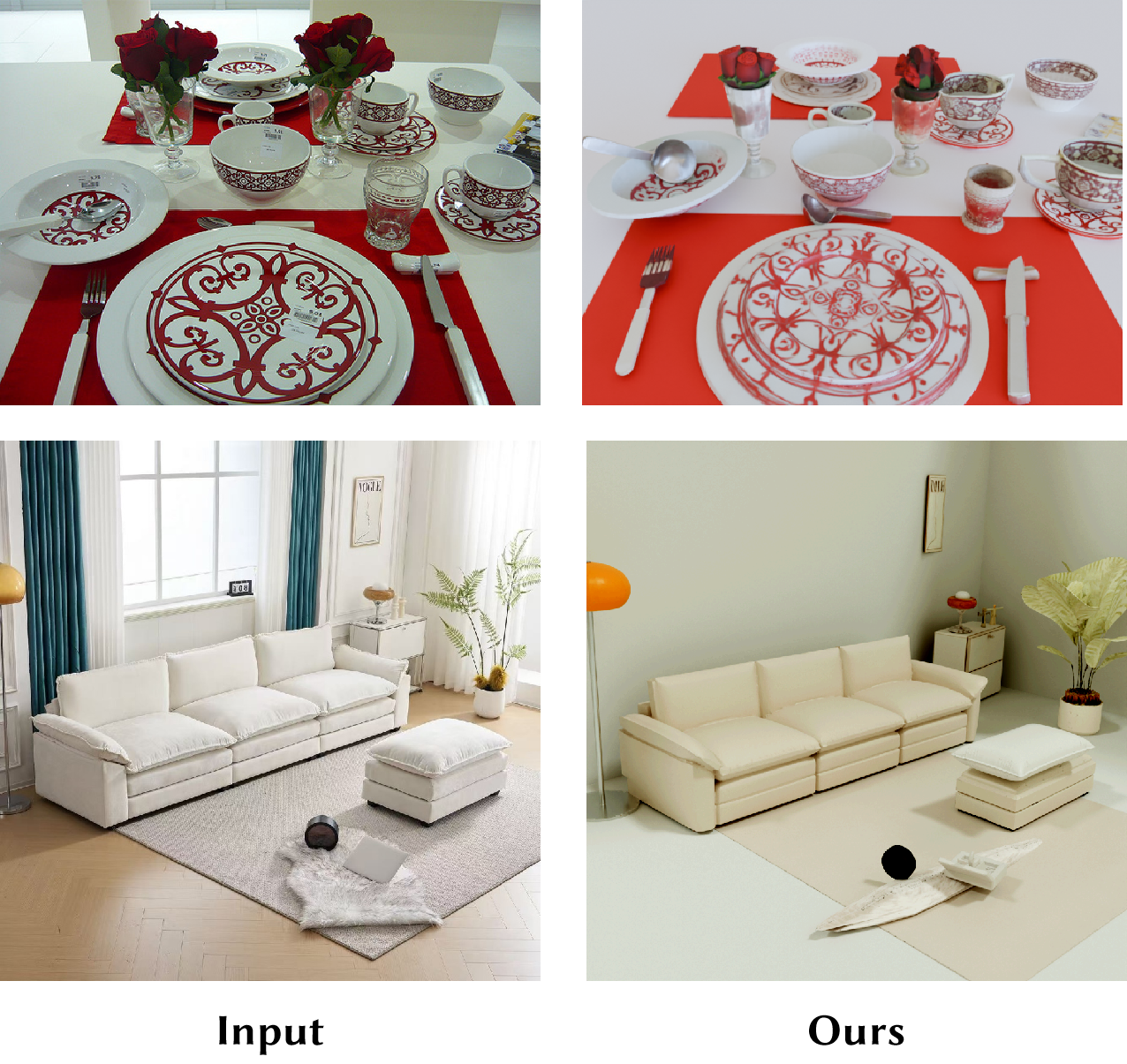}
  \captionof{figure}{In some scenes, transparent glass, textiles, and fabrics are difficult to express, as the mesh struggles to represent them realistically.} 
  \label{fig_failure}
\end{figure}

\paragraph{Limitations and Future Work}

The quality of scene generation in CAST is heavily dependent on the underlying object generation model. At present, the model still lacks sufficient detail and precision, this limitation leads to noticeable inconsistencies in the generated objects, affecting their alignment and spatial relations in the scene. Additionally, the current mesh representation struggles with materials like textiles, glasses or fabrics, often appearing unnatural, and fails to accurately depict transparent materials, as shown in Fig. \ref{fig_failure}. Although additional modules have been incorporated to enhance object robustness and similarity, the need for more advanced and robust generation models remains. A more detailed and accurate object generator could significantly improve the overall scene quality and enhance its real-world applicability.

A notable limitation of the current method is the absence of lighting estimation and background modeling. Without realistic lighting, the interactions between objects and their surroundings may lack natural shading and illumination effects, impacting the visual realism and immersion of the generated 3D environments. For enhanced visual realism, we employ an off-the-shelf panoramic HDR generation tool \cite{hyper3d-hdri} combined with preset lighting conditions in blender manually. Future enhancements in CAST could benefit from integrating advanced techniques for lighting estimation and background modeling, which would significantly enrich the contextual depth and visual fidelity of the scenes.

In more complex scenes, the performance of the current method may experience slight degradation. Challenges such as intricate spatial layouts and dense object configurations could affect the accuracy of scene reconstruction to some extent. While CAST currently excels at reconstructing individual scenes, there is significant potential to utilize its outputs to build large-scale datasets, facilitating advanced research on fully learned scene or video generation pipelines. Expanding the variety and realism of generated scenes in this manner could further improve the robustness and applicability of 3D generative models in areas such as film production, simulation, and immersive media.

%% file: sec/appendix.tex
\section{GPT-4v prompt}
The following prompt instructs GPT-4V to act as an object–relationship analyst for a numbered scene image.  It defines six relationship types (Stack, Lean, Hang, Clamped, Contained, Edge/Point), sets strict formatting rules (JSON objects in a list). For full details, see the appendix prompt as below.

\begin{lstlisting}[language=Python,
                   caption={System prompt for GPT-4V},
                   label={lst:system}]
prompting_text_system = """
You are an expert in object recognition and spatial reasoning.

### Task Description ###
Analyze an image with numbered objects and determine their relationships.
For each pair of related objects, output a JSON object containing the relationship details.
Ensure you output all possible relationships, even those that may be difficult to judge or less obvious.

### Relationship Definitions ###
1. **Stack**: Object 1 is on top of Object 2 (Object 2 supports Object 1 from below)
2. **Lean**: Object 1 is leaning against Object 2 (Object 2 supports Object 1 laterally)
3. **Hang**: Object 1 is hanging from Object 2 (Object 2 supports Object 1 from above)
4. **Clamped**: Object 1 is clamped by Object 2 (Object 2 grips Object 1 on multiple sides)
5. **Contained**: Object 1 is inside Object 2 (Object 2 encloses Object 1)
6. **Edge/Point**: Object 1 is touching Object 2 at an edge or point (minimal contact, no significant support)

### Important Note ###
Only objects that are in contact with each other should have a relationship.
For each relationship, always ensure the following:
1. Use the correct relationship type.
2. Provide a clear explanation of the relationship.
3. For cases that are in contact but hard to choose which type, use "Stack".
### Output Format ###
{
  'pair': [obj1_num, obj2_num],
  'relationship': 'Stack'/'Lean'/'Hang'/'Clamped'/'Contained'/'Edge/Point',
  'reason': 'explanation'
}

### Examples ###
{ 'pair': [1, 2],  'relationship': 'Stack',      'reason': 'Book (1) is stacked on top of Table (2)' }
{ 'pair': [3, 4],  'relationship': 'Lean',       'reason': 'A chair (3) is leaning against a wall (4)' }
{ 'pair': [5, 6],  'relationship': 'Hang',       'reason': 'A lamp (5) is hanging from the ceiling (6)' }
{ 'pair': [7, 8],  'relationship': 'Clamped',    'reason': 'A pipe (7) is clamped by a bracket (8)' }
{ 'pair': [9, 10], 'relationship': 'Contained',  'reason': 'A pencil (9) is inside a pencil case (10)' }
{ 'pair': [11,12], 'relationship': 'Edge/Point', 'reason': 'A book (11) and a pen (12) are touching at the edge' }
Only objects that are in contact with each other should have a relationship.
"""

content_user = [
    {
        "type": "text",
        "text": ('''
            "### Object Details ###"
            "I have labeled a bright numeric ID at the center for each visual object in the image."
            "Please analyze all relationships between the numbered objects and output JSON objects following the specified format. "
            "Ensure each relationship includes:"
            "1. The correct relationship type"
            "2. A clear reason for the relationship"
            "List all relationships as a JSON array."
        '''),
    },
    {
        "type": "image_url",
        "image_url": {"url": f"data:image/png;base64,{base64_image}"},
    },
]
\end{lstlisting}

%% file: main.bbl

\begin{thebibliography}{97}


\ifx \showCODEN    \undefined \def \showCODEN     #1{\unskip}     \fi
\ifx \showDOI      \undefined \def \showDOI       #1{#1}\fi
\ifx \showISBNx    \undefined \def \showISBNx     #1{\unskip}     \fi
\ifx \showISBNxiii \undefined \def \showISBNxiii  #1{\unskip}     \fi
\ifx \showISSN     \undefined \def \showISSN      #1{\unskip}     \fi
\ifx \showLCCN     \undefined \def \showLCCN      #1{\unskip}     \fi
\ifx \shownote     \undefined \def \shownote      #1{#1}          \fi
\ifx \showarticletitle \undefined \def \showarticletitle #1{#1}   \fi
\ifx \showURL      \undefined \def \showURL       {\relax}        \fi
\providecommand\bibfield[2]{#2}
\providecommand\bibinfo[2]{#2}
\providecommand\natexlab[1]{#1}
\providecommand\showeprint[2][]{arXiv:#2}

\bibitem[Achiam et~al\mbox{.}(2023)]%
        {achiam2023gpt}
\bibfield{author}{\bibinfo{person}{Josh Achiam}, \bibinfo{person}{Steven Adler}, \bibinfo{person}{Sandhini Agarwal}, \bibinfo{person}{Lama Ahmad}, \bibinfo{person}{Ilge Akkaya}, \bibinfo{person}{Florencia~Leoni Aleman}, \bibinfo{person}{Diogo Almeida}, \bibinfo{person}{Janko Altenschmidt}, \bibinfo{person}{Sam Altman}, \bibinfo{person}{Shyamal Anadkat}, {et~al\mbox{.}}} \bibinfo{year}{2023}\natexlab{}.
\newblock \showarticletitle{Gpt-4 technical report}.
\newblock \bibinfo{journal}{\emph{arXiv preprint arXiv:2303.08774}} (\bibinfo{year}{2023}).
\newblock


\bibitem[Arun et~al\mbox{.}(1987)]%
        {arun1987least}
\bibfield{author}{\bibinfo{person}{K~Somani Arun}, \bibinfo{person}{Thomas~S Huang}, {and} \bibinfo{person}{Steven~D Blostein}.} \bibinfo{year}{1987}\natexlab{}.
\newblock \showarticletitle{Least-squares fitting of two 3-D point sets}.
\newblock \bibinfo{journal}{\emph{IEEE Transactions on pattern analysis and machine intelligence}} \bibinfo{number}{5} (\bibinfo{year}{1987}), \bibinfo{pages}{698--700}.
\newblock


\bibitem[Barron et~al\mbox{.}(2021)]%
        {barron2021mip}
\bibfield{author}{\bibinfo{person}{Jonathan~T Barron}, \bibinfo{person}{Ben Mildenhall}, \bibinfo{person}{Matthew Tancik}, \bibinfo{person}{Peter Hedman}, \bibinfo{person}{Ricardo Martin-Brualla}, {and} \bibinfo{person}{Pratul~P Srinivasan}.} \bibinfo{year}{2021}\natexlab{}.
\newblock \showarticletitle{Mip-nerf: A multiscale representation for anti-aliasing neural radiance fields}. In \bibinfo{booktitle}{\emph{Proceedings of the IEEE/CVF international conference on computer vision}}. \bibinfo{pages}{5855--5864}.
\newblock


\bibitem[Barron et~al\mbox{.}(2022)]%
        {barron2022mip}
\bibfield{author}{\bibinfo{person}{Jonathan~T Barron}, \bibinfo{person}{Ben Mildenhall}, \bibinfo{person}{Dor Verbin}, \bibinfo{person}{Pratul~P Srinivasan}, {and} \bibinfo{person}{Peter Hedman}.} \bibinfo{year}{2022}\natexlab{}.
\newblock \showarticletitle{Mip-nerf 360: Unbounded anti-aliased neural radiance fields}. In \bibinfo{booktitle}{\emph{Proceedings of the IEEE/CVF Conference on Computer Vision and Pattern Recognition}}. \bibinfo{pages}{5470--5479}.
\newblock


\bibitem[Bender et~al\mbox{.}(2012)]%
        {BenderETC12}
\bibfield{author}{\bibinfo{person}{Jan Bender}, \bibinfo{person}{Kenny Erleben}, \bibinfo{person}{Jeff Trinkle}, {and} \bibinfo{person}{Erwin Coumans}.} \bibinfo{year}{2012}\natexlab{}.
\newblock \showarticletitle{Interactive Simulation of Rigid Body Dynamics in Computer Graphics}. In \bibinfo{booktitle}{\emph{33rd Annual Conference of the European Association for Computer Graphics, Eurographics 2012 - State of the Art Reports, Cagliari, Sardinia, Italy, May 13-18, 2012}}, \bibfield{editor}{\bibinfo{person}{Marie{-}Paule Cani} {and} \bibinfo{person}{Fabio Ganovelli}} (Eds.). \bibinfo{publisher}{Eurographics Association}, \bibinfo{pages}{95--134}.
\newblock
\urldef\tempurl%
\url{https://doi.org/10.2312/CONF/EG2012/STARS/095-134}
\showDOI{\tempurl}


\bibitem[Best(1992)]%
        {best1992method}
\bibfield{author}{\bibinfo{person}{Paul~J Best}.} \bibinfo{year}{1992}\natexlab{}.
\newblock \showarticletitle{A method for registration of 3-D shapes}.
\newblock \bibinfo{journal}{\emph{IEEE Trans Pattern Anal Mach Vision}}  \bibinfo{volume}{14} (\bibinfo{year}{1992}), \bibinfo{pages}{239--256}.
\newblock


\bibitem[Bhat et~al\mbox{.}(2023)]%
        {bhat2023zoedepth}
\bibfield{author}{\bibinfo{person}{Shariq~Farooq Bhat}, \bibinfo{person}{Reiner Birkl}, \bibinfo{person}{Diana Wofk}, \bibinfo{person}{Peter Wonka}, {and} \bibinfo{person}{Matthias M{\"u}ller}.} \bibinfo{year}{2023}\natexlab{}.
\newblock \showarticletitle{Zoedepth: Zero-shot transfer by combining relative and metric depth}.
\newblock \bibinfo{journal}{\emph{arXiv preprint arXiv:2302.12288}} (\bibinfo{year}{2023}).
\newblock


\bibitem[Blattmann et~al\mbox{.}(2023)]%
        {blattmann2023stable}
\bibfield{author}{\bibinfo{person}{Andreas Blattmann}, \bibinfo{person}{Tim Dockhorn}, \bibinfo{person}{Sumith Kulal}, \bibinfo{person}{Daniel Mendelevitch}, \bibinfo{person}{Maciej Kilian}, \bibinfo{person}{Dominik Lorenz}, \bibinfo{person}{Yam Levi}, \bibinfo{person}{Zion English}, \bibinfo{person}{Vikram Voleti}, \bibinfo{person}{Adam Letts}, {et~al\mbox{.}}} \bibinfo{year}{2023}\natexlab{}.
\newblock \showarticletitle{Stable video diffusion: Scaling latent video diffusion models to large datasets}.
\newblock \bibinfo{journal}{\emph{arXiv preprint arXiv:2311.15127}} (\bibinfo{year}{2023}).
\newblock


\bibitem[Bruce et~al\mbox{.}(2024)]%
        {bruce2024genie}
\bibfield{author}{\bibinfo{person}{Jake Bruce}, \bibinfo{person}{Michael~D Dennis}, \bibinfo{person}{Ashley Edwards}, \bibinfo{person}{Jack Parker-Holder}, \bibinfo{person}{Yuge Shi}, \bibinfo{person}{Edward Hughes}, \bibinfo{person}{Matthew Lai}, \bibinfo{person}{Aditi Mavalankar}, \bibinfo{person}{Richie Steigerwald}, \bibinfo{person}{Chris Apps}, {et~al\mbox{.}}} \bibinfo{year}{2024}\natexlab{}.
\newblock \showarticletitle{Genie: Generative interactive environments}. In \bibinfo{booktitle}{\emph{Forty-first International Conference on Machine Learning}}.
\newblock


\bibitem[Chang et~al\mbox{.}(2015)]%
        {chang2015shapenet}
\bibfield{author}{\bibinfo{person}{Angel~X Chang}, \bibinfo{person}{Thomas Funkhouser}, \bibinfo{person}{Leonidas Guibas}, \bibinfo{person}{Pat Hanrahan}, \bibinfo{person}{Qixing Huang}, \bibinfo{person}{Zimo Li}, \bibinfo{person}{Silvio Savarese}, \bibinfo{person}{Manolis Savva}, \bibinfo{person}{Shuran Song}, \bibinfo{person}{Hao Su}, {et~al\mbox{.}}} \bibinfo{year}{2015}\natexlab{}.
\newblock \showarticletitle{Shapenet: An information-rich 3d model repository}.
\newblock \bibinfo{journal}{\emph{arXiv preprint arXiv:1512.03012}} (\bibinfo{year}{2015}).
\newblock


\bibitem[Chen et~al\mbox{.}(2018)]%
        {chen2018deep}
\bibfield{author}{\bibinfo{person}{Anpei Chen}, \bibinfo{person}{Minye Wu}, \bibinfo{person}{Yingliang Zhang}, \bibinfo{person}{Nianyi Li}, \bibinfo{person}{Jie Lu}, \bibinfo{person}{Shenghua Gao}, {and} \bibinfo{person}{Jingyi Yu}.} \bibinfo{year}{2018}\natexlab{}.
\newblock \showarticletitle{Deep surface light fields}.
\newblock \bibinfo{journal}{\emph{Proceedings of the ACM on Computer Graphics and Interactive Techniques}} \bibinfo{volume}{1}, \bibinfo{number}{1} (\bibinfo{year}{2018}), \bibinfo{pages}{1--17}.
\newblock


\bibitem[Chen et~al\mbox{.}(2024a)]%
        {chen2024single}
\bibfield{author}{\bibinfo{person}{Yixin Chen}, \bibinfo{person}{Junfeng Ni}, \bibinfo{person}{Nan Jiang}, \bibinfo{person}{Yaowei Zhang}, \bibinfo{person}{Yixin Zhu}, {and} \bibinfo{person}{Siyuan Huang}.} \bibinfo{year}{2024}\natexlab{a}.
\newblock \showarticletitle{Single-view 3d scene reconstruction with high-fidelity shape and texture}. In \bibinfo{booktitle}{\emph{2024 International Conference on 3D Vision (3DV)}}. IEEE, \bibinfo{pages}{1456--1467}.
\newblock


\bibitem[Chen et~al\mbox{.}(2024b)]%
        {chen2024atlas3d}
\bibfield{author}{\bibinfo{person}{Yunuo Chen}, \bibinfo{person}{Tianyi Xie}, \bibinfo{person}{Zeshun Zong}, \bibinfo{person}{Xuan Li}, \bibinfo{person}{Feng Gao}, \bibinfo{person}{Yin Yang}, \bibinfo{person}{Ying~Nian Wu}, {and} \bibinfo{person}{Chenfanfu Jiang}.} \bibinfo{year}{2024}\natexlab{b}.
\newblock \showarticletitle{Atlas3D: Physically Constrained Self-Supporting Text-to-3D for Simulation and Fabrication}.
\newblock \bibinfo{journal}{\emph{arXiv preprint arXiv:2405.18515}} (\bibinfo{year}{2024}).
\newblock


\bibitem[Cheng et~al\mbox{.}(2024)]%
        {cheng2024navila}
\bibfield{author}{\bibinfo{person}{An-Chieh Cheng}, \bibinfo{person}{Yandong Ji}, \bibinfo{person}{Zhaojing Yang}, \bibinfo{person}{Xueyan Zou}, \bibinfo{person}{Jan Kautz}, \bibinfo{person}{Erdem B{\i}y{\i}k}, \bibinfo{person}{Hongxu Yin}, \bibinfo{person}{Sifei Liu}, {and} \bibinfo{person}{Xiaolong Wang}.} \bibinfo{year}{2024}\natexlab{}.
\newblock \showarticletitle{Navila: Legged robot vision-language-action model for navigation}.
\newblock \bibinfo{journal}{\emph{arXiv preprint arXiv:2412.04453}} (\bibinfo{year}{2024}).
\newblock


\bibitem[Chu et~al\mbox{.}(2023)]%
        {chu2023buol}
\bibfield{author}{\bibinfo{person}{Tao Chu}, \bibinfo{person}{Pan Zhang}, \bibinfo{person}{Qiong Liu}, {and} \bibinfo{person}{Jiaqi Wang}.} \bibinfo{year}{2023}\natexlab{}.
\newblock \showarticletitle{Buol: A bottom-up framework with occupancy-aware lifting for panoptic 3d scene reconstruction from a single image}. In \bibinfo{booktitle}{\emph{Proceedings of the IEEE/CVF Conference on Computer Vision and Pattern Recognition}}. \bibinfo{pages}{4937--4946}.
\newblock


\bibitem[Dahnert et~al\mbox{.}(2021)]%
        {dahnert2021panoptic}
\bibfield{author}{\bibinfo{person}{Manuel Dahnert}, \bibinfo{person}{Ji Hou}, \bibinfo{person}{Matthias Nie{\ss}ner}, {and} \bibinfo{person}{Angela Dai}.} \bibinfo{year}{2021}\natexlab{}.
\newblock \showarticletitle{Panoptic 3d scene reconstruction from a single rgb image}.
\newblock \bibinfo{journal}{\emph{Advances in Neural Information Processing Systems}}  \bibinfo{volume}{34} (\bibinfo{year}{2021}), \bibinfo{pages}{8282--8293}.
\newblock


\bibitem[Dai et~al\mbox{.}(2017)]%
        {dai2017scannet}
\bibfield{author}{\bibinfo{person}{Angela Dai}, \bibinfo{person}{Angel~X Chang}, \bibinfo{person}{Manolis Savva}, \bibinfo{person}{Maciej Halber}, \bibinfo{person}{Thomas Funkhouser}, {and} \bibinfo{person}{Matthias Nie{\ss}ner}.} \bibinfo{year}{2017}\natexlab{}.
\newblock \showarticletitle{Scannet: Richly-annotated 3d reconstructions of indoor scenes}. In \bibinfo{booktitle}{\emph{Proceedings of the IEEE conference on computer vision and pattern recognition}}. \bibinfo{pages}{5828--5839}.
\newblock


\bibitem[Dai et~al\mbox{.}(2024)]%
        {dai2024automated}
\bibfield{author}{\bibinfo{person}{Tianyuan Dai}, \bibinfo{person}{Josiah Wong}, \bibinfo{person}{Yunfan Jiang}, \bibinfo{person}{Chen Wang}, \bibinfo{person}{Cem Gokmen}, \bibinfo{person}{Ruohan Zhang}, \bibinfo{person}{Jiajun Wu}, {and} \bibinfo{person}{Li Fei-Fei}.} \bibinfo{year}{2024}\natexlab{}.
\newblock \showarticletitle{Automated Creation of Digital Cousins for Robust Policy Learning}.
\newblock \bibinfo{journal}{\emph{arXiv preprint arXiv:2410.07408}} (\bibinfo{year}{2024}).
\newblock


\bibitem[Deitke et~al\mbox{.}(2024)]%
        {deitke2024objaverse}
\bibfield{author}{\bibinfo{person}{Matt Deitke}, \bibinfo{person}{Ruoshi Liu}, \bibinfo{person}{Matthew Wallingford}, \bibinfo{person}{Huong Ngo}, \bibinfo{person}{Oscar Michel}, \bibinfo{person}{Aditya Kusupati}, \bibinfo{person}{Alan Fan}, \bibinfo{person}{Christian Laforte}, \bibinfo{person}{Vikram Voleti}, \bibinfo{person}{Samir~Yitzhak Gadre}, {et~al\mbox{.}}} \bibinfo{year}{2024}\natexlab{}.
\newblock \showarticletitle{Objaverse-xl: A universe of 10m+ 3d objects}.
\newblock \bibinfo{journal}{\emph{Advances in Neural Information Processing Systems}}  \bibinfo{volume}{36} (\bibinfo{year}{2024}).
\newblock


\bibitem[Deitke et~al\mbox{.}(2023)]%
        {deitke2023objaverse}
\bibfield{author}{\bibinfo{person}{Matt Deitke}, \bibinfo{person}{Dustin Schwenk}, \bibinfo{person}{Jordi Salvador}, \bibinfo{person}{Luca Weihs}, \bibinfo{person}{Oscar Michel}, \bibinfo{person}{Eli VanderBilt}, \bibinfo{person}{Ludwig Schmidt}, \bibinfo{person}{Kiana Ehsani}, \bibinfo{person}{Aniruddha Kembhavi}, {and} \bibinfo{person}{Ali Farhadi}.} \bibinfo{year}{2023}\natexlab{}.
\newblock \showarticletitle{Objaverse: A universe of annotated 3d objects}. In \bibinfo{booktitle}{\emph{Proceedings of the IEEE/CVF Conference on Computer Vision and Pattern Recognition}}. \bibinfo{pages}{13142--13153}.
\newblock


\bibitem[Dogaru et~al\mbox{.}(2024)]%
        {dogaru2024generalizable}
\bibfield{author}{\bibinfo{person}{Andreea Dogaru}, \bibinfo{person}{Mert {\"O}zer}, {and} \bibinfo{person}{Bernhard Egger}.} \bibinfo{year}{2024}\natexlab{}.
\newblock \showarticletitle{Generalizable 3D Scene Reconstruction via Divide and Conquer from a Single View}.
\newblock \bibinfo{journal}{\emph{arXiv preprint arXiv:2404.03421}} (\bibinfo{year}{2024}).
\newblock


\bibitem[Fu et~al\mbox{.}(2021)]%
        {fu20213d}
\bibfield{author}{\bibinfo{person}{Huan Fu}, \bibinfo{person}{Bowen Cai}, \bibinfo{person}{Lin Gao}, \bibinfo{person}{Ling-Xiao Zhang}, \bibinfo{person}{Jiaming Wang}, \bibinfo{person}{Cao Li}, \bibinfo{person}{Qixun Zeng}, \bibinfo{person}{Chengyue Sun}, \bibinfo{person}{Rongfei Jia}, \bibinfo{person}{Binqiang Zhao}, {et~al\mbox{.}}} \bibinfo{year}{2021}\natexlab{}.
\newblock \showarticletitle{3d-front: 3d furnished rooms with layouts and semantics}. In \bibinfo{booktitle}{\emph{Proceedings of the IEEE/CVF International Conference on Computer Vision}}. \bibinfo{pages}{10933--10942}.
\newblock


\bibitem[Gao et~al\mbox{.}(2024b)]%
        {gao2024diffcad}
\bibfield{author}{\bibinfo{person}{Daoyi Gao}, \bibinfo{person}{D{\'a}vid Rozenberszki}, \bibinfo{person}{Stefan Leutenegger}, {and} \bibinfo{person}{Angela Dai}.} \bibinfo{year}{2024}\natexlab{b}.
\newblock \showarticletitle{Diffcad: Weakly-supervised probabilistic cad model retrieval and alignment from an rgb image}.
\newblock \bibinfo{journal}{\emph{ACM Transactions on Graphics (TOG)}} \bibinfo{volume}{43}, \bibinfo{number}{4} (\bibinfo{year}{2024}), \bibinfo{pages}{1--15}.
\newblock


\bibitem[Gao et~al\mbox{.}(2024a)]%
        {gao2024cat3d}
\bibfield{author}{\bibinfo{person}{Ruiqi Gao}, \bibinfo{person}{Aleksander Holynski}, \bibinfo{person}{Philipp Henzler}, \bibinfo{person}{Arthur Brussee}, \bibinfo{person}{Ricardo Martin-Brualla}, \bibinfo{person}{Pratul Srinivasan}, \bibinfo{person}{Jonathan~T Barron}, {and} \bibinfo{person}{Ben Poole}.} \bibinfo{year}{2024}\natexlab{a}.
\newblock \showarticletitle{Cat3d: Create anything in 3d with multi-view diffusion models}.
\newblock \bibinfo{journal}{\emph{arXiv preprint arXiv:2405.10314}} (\bibinfo{year}{2024}).
\newblock


\bibitem[Geiger et~al\mbox{.}(2013)]%
        {geiger2013vision}
\bibfield{author}{\bibinfo{person}{Andreas Geiger}, \bibinfo{person}{Philip Lenz}, \bibinfo{person}{Christoph Stiller}, {and} \bibinfo{person}{Raquel Urtasun}.} \bibinfo{year}{2013}\natexlab{}.
\newblock \showarticletitle{Vision meets robotics: The kitti dataset}.
\newblock \bibinfo{journal}{\emph{The International Journal of Robotics Research}} \bibinfo{volume}{32}, \bibinfo{number}{11} (\bibinfo{year}{2013}), \bibinfo{pages}{1231--1237}.
\newblock


\bibitem[Gkioxari et~al\mbox{.}(2022)]%
        {gkioxari2022learning}
\bibfield{author}{\bibinfo{person}{Georgia Gkioxari}, \bibinfo{person}{Nikhila Ravi}, {and} \bibinfo{person}{Justin Johnson}.} \bibinfo{year}{2022}\natexlab{}.
\newblock \showarticletitle{Learning 3d object shape and layout without 3d supervision}. In \bibinfo{booktitle}{\emph{Proceedings of the IEEE/CVF Conference on Computer Vision and Pattern Recognition}}. \bibinfo{pages}{1695--1704}.
\newblock


\bibitem[Goesele et~al\mbox{.}(2007)]%
        {MVS}
\bibfield{author}{\bibinfo{person}{Michael Goesele}, \bibinfo{person}{Noah Snavely}, \bibinfo{person}{Brian Curless}, \bibinfo{person}{Hugues Hoppe}, {and} \bibinfo{person}{Steven~M. Seitz}.} \bibinfo{year}{2007}\natexlab{}.
\newblock \showarticletitle{Multi-View Stereo for Community Photo Collections}. In \bibinfo{booktitle}{\emph{2007 IEEE 11th International Conference on Computer Vision}}. \bibinfo{pages}{1--8}.
\newblock
\urldef\tempurl%
\url{https://doi.org/10.1109/ICCV.2007.4408933}
\showDOI{\tempurl}


\bibitem[G{\"u}meli et~al\mbox{.}(2022)]%
        {gumeli2022roca}
\bibfield{author}{\bibinfo{person}{Can G{\"u}meli}, \bibinfo{person}{Angela Dai}, {and} \bibinfo{person}{Matthias Nie{\ss}ner}.} \bibinfo{year}{2022}\natexlab{}.
\newblock \showarticletitle{Roca: Robust cad model retrieval and alignment from a single image}. In \bibinfo{booktitle}{\emph{Proceedings of the IEEE/CVF conference on computer vision and pattern recognition}}. \bibinfo{pages}{4022--4031}.
\newblock


\bibitem[Guo et~al\mbox{.}(2024)]%
        {guo2024physically}
\bibfield{author}{\bibinfo{person}{Minghao Guo}, \bibinfo{person}{Bohan Wang}, \bibinfo{person}{Pingchuan Ma}, \bibinfo{person}{Tianyuan Zhang}, \bibinfo{person}{Crystal~Elaine Owens}, \bibinfo{person}{Chuang Gan}, \bibinfo{person}{Joshua~B Tenenbaum}, \bibinfo{person}{Kaiming He}, {and} \bibinfo{person}{Wojciech Matusik}.} \bibinfo{year}{2024}\natexlab{}.
\newblock \showarticletitle{Physically Compatible 3D Object Modeling from a Single Image}.
\newblock \bibinfo{journal}{\emph{arXiv preprint arXiv:2405.20510}} (\bibinfo{year}{2024}).
\newblock


\bibitem[Ho et~al\mbox{.}(2022a)]%
        {ho2022imagen}
\bibfield{author}{\bibinfo{person}{Jonathan Ho}, \bibinfo{person}{William Chan}, \bibinfo{person}{Chitwan Saharia}, \bibinfo{person}{Jay Whang}, \bibinfo{person}{Ruiqi Gao}, \bibinfo{person}{Alexey Gritsenko}, \bibinfo{person}{Diederik~P Kingma}, \bibinfo{person}{Ben Poole}, \bibinfo{person}{Mohammad Norouzi}, \bibinfo{person}{David~J Fleet}, {et~al\mbox{.}}} \bibinfo{year}{2022}\natexlab{a}.
\newblock \showarticletitle{Imagen video: High definition video generation with diffusion models}.
\newblock \bibinfo{journal}{\emph{arXiv preprint arXiv:2210.02303}} (\bibinfo{year}{2022}).
\newblock


\bibitem[Ho et~al\mbox{.}(2022b)]%
        {ho2022video}
\bibfield{author}{\bibinfo{person}{Jonathan Ho}, \bibinfo{person}{Tim Salimans}, \bibinfo{person}{Alexey Gritsenko}, \bibinfo{person}{William Chan}, \bibinfo{person}{Mohammad Norouzi}, {and} \bibinfo{person}{David~J Fleet}.} \bibinfo{year}{2022}\natexlab{b}.
\newblock \showarticletitle{Video diffusion models}.
\newblock \bibinfo{journal}{\emph{Advances in Neural Information Processing Systems}}  \bibinfo{volume}{35} (\bibinfo{year}{2022}), \bibinfo{pages}{8633--8646}.
\newblock


\bibitem[Hong et~al\mbox{.}(2023)]%
        {hong2023lrm}
\bibfield{author}{\bibinfo{person}{Yicong Hong}, \bibinfo{person}{Kai Zhang}, \bibinfo{person}{Jiuxiang Gu}, \bibinfo{person}{Sai Bi}, \bibinfo{person}{Yang Zhou}, \bibinfo{person}{Difan Liu}, \bibinfo{person}{Feng Liu}, \bibinfo{person}{Kalyan Sunkavalli}, \bibinfo{person}{Trung Bui}, {and} \bibinfo{person}{Hao Tan}.} \bibinfo{year}{2023}\natexlab{}.
\newblock \showarticletitle{Lrm: Large reconstruction model for single image to 3d}.
\newblock \bibinfo{journal}{\emph{arXiv preprint arXiv:2311.04400}} (\bibinfo{year}{2023}).
\newblock


\bibitem[Huang et~al\mbox{.}(2024)]%
        {huang2024midi}
\bibfield{author}{\bibinfo{person}{Zehuan Huang}, \bibinfo{person}{Yuan-Chen Guo}, \bibinfo{person}{Xingqiao An}, \bibinfo{person}{Yunhan Yang}, \bibinfo{person}{Yangguang Li}, \bibinfo{person}{Zi-Xin Zou}, \bibinfo{person}{Ding Liang}, \bibinfo{person}{Xihui Liu}, \bibinfo{person}{Yan-Pei Cao}, {and} \bibinfo{person}{Lu Sheng}.} \bibinfo{year}{2024}\natexlab{}.
\newblock \showarticletitle{MIDI: Multi-Instance Diffusion for Single Image to 3D Scene Generation}.
\newblock \bibinfo{journal}{\emph{arXiv preprint arXiv:2412.03558}} (\bibinfo{year}{2024}).
\newblock


\bibitem[{Hyper3D}(2025)]%
        {hyper3d-hdri}
\bibfield{author}{\bibinfo{person}{{Hyper3D}}.} \bibinfo{year}{2025}\natexlab{}.
\newblock \bibinfo{booktitle}{\emph{Omnicraft}}.
\newblock
\urldef\tempurl%
\url{https://hyper3d.ai/omnicraft/hdri}
\showURL{%
\tempurl}


\bibitem[Kehl et~al\mbox{.}(2017)]%
        {kehl2017ssd}
\bibfield{author}{\bibinfo{person}{Wadim Kehl}, \bibinfo{person}{Fabian Manhardt}, \bibinfo{person}{Federico Tombari}, \bibinfo{person}{Slobodan Ilic}, {and} \bibinfo{person}{Nassir Navab}.} \bibinfo{year}{2017}\natexlab{}.
\newblock \showarticletitle{Ssd-6d: Making rgb-based 3d detection and 6d pose estimation great again}. In \bibinfo{booktitle}{\emph{Proceedings of the IEEE international conference on computer vision}}. \bibinfo{pages}{1521--1529}.
\newblock


\bibitem[Kerbl et~al\mbox{.}(2023)]%
        {kerbl3Dgaussians}
\bibfield{author}{\bibinfo{person}{Bernhard Kerbl}, \bibinfo{person}{Georgios Kopanas}, \bibinfo{person}{Thomas Leimk{\"u}hler}, {and} \bibinfo{person}{George Drettakis}.} \bibinfo{year}{2023}\natexlab{}.
\newblock \showarticletitle{3D Gaussian Splatting for Real-Time Radiance Field Rendering}.
\newblock \bibinfo{journal}{\emph{ACM Transactions on Graphics}} \bibinfo{volume}{42}, \bibinfo{number}{4} (\bibinfo{date}{July} \bibinfo{year}{2023}).
\newblock
\urldef\tempurl%
\url{https://repo-sam.inria.fr/fungraph/3d-gaussian-splatting/}
\showURL{%
\tempurl}


\bibitem[Kirillov et~al\mbox{.}(2023)]%
        {kirillov2023segment}
\bibfield{author}{\bibinfo{person}{Alexander Kirillov}, \bibinfo{person}{Eric Mintun}, \bibinfo{person}{Nikhila Ravi}, \bibinfo{person}{Hanzi Mao}, \bibinfo{person}{Chloe Rolland}, \bibinfo{person}{Laura Gustafson}, \bibinfo{person}{Tete Xiao}, \bibinfo{person}{Spencer Whitehead}, \bibinfo{person}{Alexander~C Berg}, \bibinfo{person}{Wan-Yen Lo}, {et~al\mbox{.}}} \bibinfo{year}{2023}\natexlab{}.
\newblock \showarticletitle{Segment anything}. In \bibinfo{booktitle}{\emph{Proceedings of the IEEE/CVF International Conference on Computer Vision}}. \bibinfo{pages}{4015--4026}.
\newblock


\bibitem[Kuo et~al\mbox{.}(2021)]%
        {kuo2021patch2cad}
\bibfield{author}{\bibinfo{person}{Weicheng Kuo}, \bibinfo{person}{Anelia Angelova}, \bibinfo{person}{Tsung-Yi Lin}, {and} \bibinfo{person}{Angela Dai}.} \bibinfo{year}{2021}\natexlab{}.
\newblock \showarticletitle{Patch2cad: Patchwise embedding learning for in-the-wild shape retrieval from a single image}. In \bibinfo{booktitle}{\emph{Proceedings of the IEEE/CVF International Conference on Computer Vision}}. \bibinfo{pages}{12589--12599}.
\newblock


\bibitem[Labb{\'e} et~al\mbox{.}(2020)]%
        {labbe2020cosypose}
\bibfield{author}{\bibinfo{person}{Yann Labb{\'e}}, \bibinfo{person}{Justin Carpentier}, \bibinfo{person}{Mathieu Aubry}, {and} \bibinfo{person}{Josef Sivic}.} \bibinfo{year}{2020}\natexlab{}.
\newblock \showarticletitle{Cosypose: Consistent multi-view multi-object 6d pose estimation}. In \bibinfo{booktitle}{\emph{Computer Vision--ECCV 2020: 16th European Conference, Glasgow, UK, August 23--28, 2020, Proceedings, Part XVII 16}}. Springer, \bibinfo{pages}{574--591}.
\newblock


\bibitem[Laine et~al\mbox{.}(2020)]%
        {nvdiffrast}
\bibfield{author}{\bibinfo{person}{Samuli Laine}, \bibinfo{person}{Janne Hellsten}, \bibinfo{person}{Tero Karras}, \bibinfo{person}{Yeongho Seol}, \bibinfo{person}{Jaakko Lehtinen}, {and} \bibinfo{person}{Timo Aila}.} \bibinfo{year}{2020}\natexlab{}.
\newblock \showarticletitle{Modular primitives for high-performance differentiable rendering}.
\newblock \bibinfo{journal}{\emph{ACM Transactions on Graphics (TOG)}}  \bibinfo{volume}{39} (\bibinfo{year}{2020}), \bibinfo{pages}{1 -- 14}.
\newblock


\bibitem[Langer et~al\mbox{.}(2022)]%
        {langer2022sparc}
\bibfield{author}{\bibinfo{person}{Florian Langer}, \bibinfo{person}{Gwangbin Bae}, \bibinfo{person}{Ignas Budvytis}, {and} \bibinfo{person}{Roberto Cipolla}.} \bibinfo{year}{2022}\natexlab{}.
\newblock \showarticletitle{SPARC: Sparse render-and-compare for CAD model alignment in a single RGB image}.
\newblock \bibinfo{journal}{\emph{arXiv preprint arXiv:2210.01044}} (\bibinfo{year}{2022}).
\newblock


\bibitem[Latour(2005)]%
        {latour2005reassembling}
\bibfield{author}{\bibinfo{person}{Bruno Latour}.} \bibinfo{year}{2005}\natexlab{}.
\newblock \bibinfo{booktitle}{\emph{Reassembling the Social: An Introduction to Actor-Network-Theory}}.
\newblock \bibinfo{publisher}{Oxford University Press}, \bibinfo{address}{Oxford, UK}.
\newblock


\bibitem[Li et~al\mbox{.}(2024b)]%
        {li2024llm}
\bibfield{author}{\bibinfo{person}{Wenhao Li}, \bibinfo{person}{Zhiyuan Yu}, \bibinfo{person}{Qijin She}, \bibinfo{person}{Zhinan Yu}, \bibinfo{person}{Yuqing Lan}, \bibinfo{person}{Chenyang Zhu}, \bibinfo{person}{Ruizhen Hu}, {and} \bibinfo{person}{Kai Xu}.} \bibinfo{year}{2024}\natexlab{b}.
\newblock \showarticletitle{LLM-enhanced Scene Graph Learning for Household Rearrangement}. In \bibinfo{booktitle}{\emph{SIGGRAPH Asia 2024 Conference Papers}}. \bibinfo{pages}{1--11}.
\newblock


\bibitem[Li et~al\mbox{.}(2024a)]%
        {li2024evaluating}
\bibfield{author}{\bibinfo{person}{Xuanlin Li}, \bibinfo{person}{Kyle Hsu}, \bibinfo{person}{Jiayuan Gu}, \bibinfo{person}{Karl Pertsch}, \bibinfo{person}{Oier Mees}, \bibinfo{person}{Homer~Rich Walke}, \bibinfo{person}{Chuyuan Fu}, \bibinfo{person}{Ishikaa Lunawat}, \bibinfo{person}{Isabel Sieh}, \bibinfo{person}{Sean Kirmani}, {et~al\mbox{.}}} \bibinfo{year}{2024}\natexlab{a}.
\newblock \showarticletitle{Evaluating Real-World Robot Manipulation Policies in Simulation}.
\newblock \bibinfo{journal}{\emph{arXiv preprint arXiv:2405.05941}} (\bibinfo{year}{2024}).
\newblock


\bibitem[Liang et~al\mbox{.}(2024)]%
        {liang2024luciddreamer}
\bibfield{author}{\bibinfo{person}{Yixun Liang}, \bibinfo{person}{Xin Yang}, \bibinfo{person}{Jiantao Lin}, \bibinfo{person}{Haodong Li}, \bibinfo{person}{Xiaogang Xu}, {and} \bibinfo{person}{Yingcong Chen}.} \bibinfo{year}{2024}\natexlab{}.
\newblock \showarticletitle{Luciddreamer: Towards high-fidelity text-to-3d generation via interval score matching}. In \bibinfo{booktitle}{\emph{Proceedings of the IEEE/CVF Conference on Computer Vision and Pattern Recognition}}. \bibinfo{pages}{6517--6526}.
\newblock


\bibitem[Liu et~al\mbox{.}(2022)]%
        {liu2022towards}
\bibfield{author}{\bibinfo{person}{Haolin Liu}, \bibinfo{person}{Yujian Zheng}, \bibinfo{person}{Guanying Chen}, \bibinfo{person}{Shuguang Cui}, {and} \bibinfo{person}{Xiaoguang Han}.} \bibinfo{year}{2022}\natexlab{}.
\newblock \showarticletitle{Towards high-fidelity single-view holistic reconstruction of indoor scenes}. In \bibinfo{booktitle}{\emph{European Conference on Computer Vision}}. Springer, \bibinfo{pages}{429--446}.
\newblock


\bibitem[Liu et~al\mbox{.}(2024)]%
        {liu2024one}
\bibfield{author}{\bibinfo{person}{Minghua Liu}, \bibinfo{person}{Chao Xu}, \bibinfo{person}{Haian Jin}, \bibinfo{person}{Linghao Chen}, \bibinfo{person}{Mukund Varma~T}, \bibinfo{person}{Zexiang Xu}, {and} \bibinfo{person}{Hao Su}.} \bibinfo{year}{2024}\natexlab{}.
\newblock \showarticletitle{One-2-3-45: Any single image to 3d mesh in 45 seconds without per-shape optimization}.
\newblock \bibinfo{journal}{\emph{Advances in Neural Information Processing Systems}}  \bibinfo{volume}{36} (\bibinfo{year}{2024}).
\newblock


\bibitem[Liu et~al\mbox{.}(2023c)]%
        {liu2023zero}
\bibfield{author}{\bibinfo{person}{Ruoshi Liu}, \bibinfo{person}{Rundi Wu}, \bibinfo{person}{Basile Van~Hoorick}, \bibinfo{person}{Pavel Tokmakov}, \bibinfo{person}{Sergey Zakharov}, {and} \bibinfo{person}{Carl Vondrick}.} \bibinfo{year}{2023}\natexlab{c}.
\newblock \showarticletitle{Zero-1-to-3: Zero-shot one image to 3d object}. In \bibinfo{booktitle}{\emph{Proceedings of the IEEE/CVF International Conference on Computer Vision}}.
\newblock


\bibitem[Liu et~al\mbox{.}(2025)]%
        {liu2025grounding}
\bibfield{author}{\bibinfo{person}{Shilong Liu}, \bibinfo{person}{Zhaoyang Zeng}, \bibinfo{person}{Tianhe Ren}, \bibinfo{person}{Feng Li}, \bibinfo{person}{Hao Zhang}, \bibinfo{person}{Jie Yang}, \bibinfo{person}{Qing Jiang}, \bibinfo{person}{Chunyuan Li}, \bibinfo{person}{Jianwei Yang}, \bibinfo{person}{Hang Su}, {et~al\mbox{.}}} \bibinfo{year}{2025}\natexlab{}.
\newblock \showarticletitle{Grounding dino: Marrying dino with grounded pre-training for open-set object detection}. In \bibinfo{booktitle}{\emph{European Conference on Computer Vision}}. Springer, \bibinfo{pages}{38--55}.
\newblock


\bibitem[Liu et~al\mbox{.}(2023b)]%
        {liu2023few}
\bibfield{author}{\bibinfo{person}{Xueyi Liu}, \bibinfo{person}{Bin Wang}, \bibinfo{person}{He Wang}, {and} \bibinfo{person}{Li Yi}.} \bibinfo{year}{2023}\natexlab{b}.
\newblock \showarticletitle{Few-Shot Physically-Aware Articulated Mesh Generation via Hierarchical Deformation}. In \bibinfo{booktitle}{\emph{Proceedings of the IEEE/CVF International Conference on Computer Vision}}. \bibinfo{pages}{854--864}.
\newblock


\bibitem[Liu et~al\mbox{.}(2023a)]%
        {liu2023syncdreamer}
\bibfield{author}{\bibinfo{person}{Yuan Liu}, \bibinfo{person}{Cheng Lin}, \bibinfo{person}{Zijiao Zeng}, \bibinfo{person}{Xiaoxiao Long}, \bibinfo{person}{Lingjie Liu}, \bibinfo{person}{Taku Komura}, {and} \bibinfo{person}{Wenping Wang}.} \bibinfo{year}{2023}\natexlab{a}.
\newblock \showarticletitle{SyncDreamer: Generating Multiview-consistent Images from a Single-view Image}. In \bibinfo{booktitle}{\emph{arXiv preprint arXiv:2309.03453}}.
\newblock


\bibitem[Long et~al\mbox{.}(2024)]%
        {long2024wonder3d}
\bibfield{author}{\bibinfo{person}{Xiaoxiao Long}, \bibinfo{person}{Yuan-Chen Guo}, \bibinfo{person}{Cheng Lin}, \bibinfo{person}{Yuan Liu}, \bibinfo{person}{Zhiyang Dou}, \bibinfo{person}{Lingjie Liu}, \bibinfo{person}{Yuexin Ma}, \bibinfo{person}{Song-Hai Zhang}, \bibinfo{person}{Marc Habermann}, \bibinfo{person}{Christian Theobalt}, {et~al\mbox{.}}} \bibinfo{year}{2024}\natexlab{}.
\newblock \showarticletitle{Wonder3d: Single image to 3d using cross-domain diffusion}. In \bibinfo{booktitle}{\emph{Proceedings of the IEEE/CVF Conference on Computer Vision and Pattern Recognition}}. \bibinfo{pages}{9970--9980}.
\newblock


\bibitem[Mamou and Ghorbel(2009)]%
        {mamou2009simple}
\bibfield{author}{\bibinfo{person}{Khaled Mamou} {and} \bibinfo{person}{Faouzi Ghorbel}.} \bibinfo{year}{2009}\natexlab{}.
\newblock \showarticletitle{A simple and efficient approach for 3D mesh approximate convex decomposition}. In \bibinfo{booktitle}{\emph{2009 16th IEEE international conference on image processing (ICIP)}}. IEEE, \bibinfo{pages}{3501--3504}.
\newblock


\bibitem[Mamou et~al\mbox{.}(2016)]%
        {mamou2016volumetric}
\bibfield{author}{\bibinfo{person}{Khaled Mamou}, \bibinfo{person}{E Lengyel}, {and} \bibinfo{person}{A Peters}.} \bibinfo{year}{2016}\natexlab{}.
\newblock \showarticletitle{Volumetric hierarchical approximate convex decomposition}.
\newblock \bibinfo{journal}{\emph{Game engine gems}}  \bibinfo{volume}{3} (\bibinfo{year}{2016}), \bibinfo{pages}{141--158}.
\newblock


\bibitem[Mezghanni et~al\mbox{.}(2022)]%
        {mezghanni2022physical}
\bibfield{author}{\bibinfo{person}{Mariem Mezghanni}, \bibinfo{person}{Th{\'e}o Bodrito}, \bibinfo{person}{Malika Boulkenafed}, {and} \bibinfo{person}{Maks Ovsjanikov}.} \bibinfo{year}{2022}\natexlab{}.
\newblock \showarticletitle{Physical simulation layer for accurate 3d modeling}. In \bibinfo{booktitle}{\emph{Proceedings of the IEEE/CVF Conference on Computer Vision and Pattern Recognition}}. \bibinfo{pages}{13514--13523}.
\newblock


\bibitem[Mezghanni et~al\mbox{.}(2021)]%
        {mezghanni2021physically}
\bibfield{author}{\bibinfo{person}{Mariem Mezghanni}, \bibinfo{person}{Malika Boulkenafed}, \bibinfo{person}{Andre Lieutier}, {and} \bibinfo{person}{Maks Ovsjanikov}.} \bibinfo{year}{2021}\natexlab{}.
\newblock \showarticletitle{Physically-aware generative network for 3d shape modeling}. In \bibinfo{booktitle}{\emph{Proceedings of the IEEE/CVF Conference on Computer Vision and Pattern Recognition}}. \bibinfo{pages}{9330--9341}.
\newblock


\bibitem[Mildenhall et~al\mbox{.}(2020)]%
        {mildenhall2020nerf}
\bibfield{author}{\bibinfo{person}{Ben Mildenhall}, \bibinfo{person}{Pratul~P Srinivasan}, \bibinfo{person}{Matthew Tancik}, \bibinfo{person}{Jonathan~T Barron}, \bibinfo{person}{Ravi Ramamoorthi}, {and} \bibinfo{person}{Ren Ng}.} \bibinfo{year}{2020}\natexlab{}.
\newblock \showarticletitle{Nerf: Representing scenes as neural radiance fields for view synthesis}. In \bibinfo{booktitle}{\emph{European conference on computer vision}}. Springer, \bibinfo{pages}{405--421}.
\newblock


\bibitem[M\"uller et~al\mbox{.}(2022)]%
        {muller2022instant}
\bibfield{author}{\bibinfo{person}{Thomas M\"uller}, \bibinfo{person}{Alex Evans}, \bibinfo{person}{Christoph Schied}, {and} \bibinfo{person}{Alexander Keller}.} \bibinfo{year}{2022}\natexlab{}.
\newblock \showarticletitle{Instant Neural Graphics Primitives with a Multiresolution Hash Encoding}.
\newblock \bibinfo{journal}{\emph{ACM Trans. Graph.}} \bibinfo{volume}{41}, \bibinfo{number}{4}, Article \bibinfo{articleno}{102} (\bibinfo{date}{July} \bibinfo{year}{2022}), \bibinfo{numpages}{15}~pages.
\newblock
\urldef\tempurl%
\url{https://doi.org/10.1145/3528223.3530127}
\showDOI{\tempurl}


\bibitem[Ni et~al\mbox{.}(2024)]%
        {ni2024phyrecon}
\bibfield{author}{\bibinfo{person}{Junfeng Ni}, \bibinfo{person}{Yixin Chen}, \bibinfo{person}{Bohan Jing}, \bibinfo{person}{Nan Jiang}, \bibinfo{person}{Bin Wang}, \bibinfo{person}{Bo Dai}, \bibinfo{person}{Puhao Li}, \bibinfo{person}{Yixin Zhu}, \bibinfo{person}{Song-Chun Zhu}, {and} \bibinfo{person}{Siyuan Huang}.} \bibinfo{year}{2024}\natexlab{}.
\newblock \showarticletitle{PhyRecon: Physically Plausible Neural Scene Reconstruction}.
\newblock \bibinfo{journal}{\emph{Advances in Neural Information Processing Systems}}.
\newblock


\bibitem[Oquab et~al\mbox{.}(2023)]%
        {oquab2023dinov2}
\bibfield{author}{\bibinfo{person}{Maxime Oquab}, \bibinfo{person}{Timoth{\'e}e Darcet}, \bibinfo{person}{Th{\'e}o Moutakanni}, \bibinfo{person}{Huy Vo}, \bibinfo{person}{Marc Szafraniec}, \bibinfo{person}{Vasil Khalidov}, \bibinfo{person}{Pierre Fernandez}, \bibinfo{person}{Daniel Haziza}, \bibinfo{person}{Francisco Massa}, \bibinfo{person}{Alaaeldin El-Nouby}, {et~al\mbox{.}}} \bibinfo{year}{2023}\natexlab{}.
\newblock \showarticletitle{Dinov2: Learning robust visual features without supervision}.
\newblock \bibinfo{journal}{\emph{arXiv preprint arXiv:2304.07193}} (\bibinfo{year}{2023}).
\newblock


\bibitem[Piccinelli et~al\mbox{.}(2024)]%
        {piccinelli2024unidepth}
\bibfield{author}{\bibinfo{person}{Luigi Piccinelli}, \bibinfo{person}{Yung-Hsu Yang}, \bibinfo{person}{Christos Sakaridis}, \bibinfo{person}{Mattia Segu}, \bibinfo{person}{Siyuan Li}, \bibinfo{person}{Luc Van~Gool}, {and} \bibinfo{person}{Fisher Yu}.} \bibinfo{year}{2024}\natexlab{}.
\newblock \showarticletitle{UniDepth: Universal Monocular Metric Depth Estimation}. In \bibinfo{booktitle}{\emph{Proceedings of the IEEE/CVF Conference on Computer Vision and Pattern Recognition}}. \bibinfo{pages}{10106--10116}.
\newblock


\bibitem[Poole et~al\mbox{.}(2022)]%
        {poole2022dreamfusion}
\bibfield{author}{\bibinfo{person}{Ben Poole}, \bibinfo{person}{Ajay Jain}, \bibinfo{person}{Jonathan~T Barron}, {and} \bibinfo{person}{Ben Mildenhall}.} \bibinfo{year}{2022}\natexlab{}.
\newblock \showarticletitle{Dreamfusion: Text-to-3d using 2d diffusion}. In \bibinfo{booktitle}{\emph{arXiv preprint arXiv:2209.14988}}.
\newblock


\bibitem[Rana et~al\mbox{.}(2023)]%
        {rana2023sayplan}
\bibfield{author}{\bibinfo{person}{Krishan Rana}, \bibinfo{person}{Jesse Haviland}, \bibinfo{person}{Sourav Garg}, \bibinfo{person}{Jad Abou-Chakra}, \bibinfo{person}{Ian Reid}, {and} \bibinfo{person}{Niko Suenderhauf}.} \bibinfo{year}{2023}\natexlab{}.
\newblock \showarticletitle{Sayplan: Grounding large language models using 3d scene graphs for scalable robot task planning}. In \bibinfo{booktitle}{\emph{7th Annual Conference on Robot Learning}}.
\newblock


\bibitem[Ravi et~al\mbox{.}(2024)]%
        {ravi2024sam}
\bibfield{author}{\bibinfo{person}{Nikhila Ravi}, \bibinfo{person}{Valentin Gabeur}, \bibinfo{person}{Yuan-Ting Hu}, \bibinfo{person}{Ronghang Hu}, \bibinfo{person}{Chaitanya Ryali}, \bibinfo{person}{Tengyu Ma}, \bibinfo{person}{Haitham Khedr}, \bibinfo{person}{Roman R{\"a}dle}, \bibinfo{person}{Chloe Rolland}, \bibinfo{person}{Laura Gustafson}, {et~al\mbox{.}}} \bibinfo{year}{2024}\natexlab{}.
\newblock \showarticletitle{Sam 2: Segment anything in images and videos}.
\newblock \bibinfo{journal}{\emph{arXiv preprint arXiv:2408.00714}} (\bibinfo{year}{2024}).
\newblock


\bibitem[Ren et~al\mbox{.}(2024)]%
        {ren2024grounded}
\bibfield{author}{\bibinfo{person}{Tianhe Ren}, \bibinfo{person}{Shilong Liu}, \bibinfo{person}{Ailing Zeng}, \bibinfo{person}{Jing Lin}, \bibinfo{person}{Kunchang Li}, \bibinfo{person}{He Cao}, \bibinfo{person}{Jiayu Chen}, \bibinfo{person}{Xinyu Huang}, \bibinfo{person}{Yukang Chen}, \bibinfo{person}{Feng Yan}, {et~al\mbox{.}}} \bibinfo{year}{2024}\natexlab{}.
\newblock \showarticletitle{Grounded sam: Assembling open-world models for diverse visual tasks}.
\newblock \bibinfo{journal}{\emph{arXiv preprint arXiv:2401.14159}} (\bibinfo{year}{2024}).
\newblock


\bibitem[Sun et~al\mbox{.}(2018)]%
        {sun2018pix3d}
\bibfield{author}{\bibinfo{person}{Xingyuan Sun}, \bibinfo{person}{Jiajun Wu}, \bibinfo{person}{Xiuming Zhang}, \bibinfo{person}{Zhoutong Zhang}, \bibinfo{person}{Chengkai Zhang}, \bibinfo{person}{Tianfan Xue}, \bibinfo{person}{Joshua~B Tenenbaum}, {and} \bibinfo{person}{William~T Freeman}.} \bibinfo{year}{2018}\natexlab{}.
\newblock \showarticletitle{Pix3d: Dataset and methods for single-image 3d shape modeling}. In \bibinfo{booktitle}{\emph{Proceedings of the IEEE conference on computer vision and pattern recognition}}. \bibinfo{pages}{2974--2983}.
\newblock


\bibitem[Szymanowicz et~al\mbox{.}(2024a)]%
        {szymanowicz2024flash3d}
\bibfield{author}{\bibinfo{person}{Stanislaw Szymanowicz}, \bibinfo{person}{Eldar Insafutdinov}, \bibinfo{person}{Chuanxia Zheng}, \bibinfo{person}{Dylan Campbell}, \bibinfo{person}{Jo{\~a}o~F Henriques}, \bibinfo{person}{Christian Rupprecht}, {and} \bibinfo{person}{Andrea Vedaldi}.} \bibinfo{year}{2024}\natexlab{a}.
\newblock \showarticletitle{Flash3D: Feed-Forward Generalisable 3D Scene Reconstruction from a Single Image}.
\newblock \bibinfo{journal}{\emph{arXiv preprint arXiv:2406.04343}} (\bibinfo{year}{2024}).
\newblock


\bibitem[Szymanowicz et~al\mbox{.}(2024b)]%
        {szymanowicz2024splatter}
\bibfield{author}{\bibinfo{person}{Stanislaw Szymanowicz}, \bibinfo{person}{Chrisitian Rupprecht}, {and} \bibinfo{person}{Andrea Vedaldi}.} \bibinfo{year}{2024}\natexlab{b}.
\newblock \showarticletitle{Splatter image: Ultra-fast single-view 3d reconstruction}. In \bibinfo{booktitle}{\emph{Proceedings of the IEEE/CVF Conference on Computer Vision and Pattern Recognition}}. \bibinfo{pages}{10208--10217}.
\newblock


\bibitem[Tang et~al\mbox{.}(2025)]%
        {tang2025lgm}
\bibfield{author}{\bibinfo{person}{Jiaxiang Tang}, \bibinfo{person}{Zhaoxi Chen}, \bibinfo{person}{Xiaokang Chen}, \bibinfo{person}{Tengfei Wang}, \bibinfo{person}{Gang Zeng}, {and} \bibinfo{person}{Ziwei Liu}.} \bibinfo{year}{2025}\natexlab{}.
\newblock \showarticletitle{Lgm: Large multi-view gaussian model for high-resolution 3d content creation}. In \bibinfo{booktitle}{\emph{European Conference on Computer Vision}}. Springer, \bibinfo{pages}{1--18}.
\newblock


\bibitem[Tang et~al\mbox{.}(2023)]%
        {tang2023dreamgaussian}
\bibfield{author}{\bibinfo{person}{Jiaxiang Tang}, \bibinfo{person}{Jiawei Ren}, \bibinfo{person}{Hang Zhou}, \bibinfo{person}{Ziwei Liu}, {and} \bibinfo{person}{Gang Zeng}.} \bibinfo{year}{2023}\natexlab{}.
\newblock \showarticletitle{Dreamgaussian: Generative gaussian splatting for efficient 3d content creation}.
\newblock \bibinfo{journal}{\emph{arXiv preprint arXiv:2309.16653}} (\bibinfo{year}{2023}).
\newblock


\bibitem[Tian et~al\mbox{.}(2023)]%
        {tian2023mononerf}
\bibfield{author}{\bibinfo{person}{Fengrui Tian}, \bibinfo{person}{Shaoyi Du}, {and} \bibinfo{person}{Yueqi Duan}.} \bibinfo{year}{2023}\natexlab{}.
\newblock \showarticletitle{Mononerf: Learning a generalizable dynamic radiance field from monocular videos}. In \bibinfo{booktitle}{\emph{Proceedings of the IEEE/CVF International Conference on Computer Vision}}. \bibinfo{pages}{17903--17913}.
\newblock


\bibitem[Torne et~al\mbox{.}(2024)]%
        {torne2024reconciling}
\bibfield{author}{\bibinfo{person}{Marcel Torne}, \bibinfo{person}{Anthony Simeonov}, \bibinfo{person}{Zechu Li}, \bibinfo{person}{April Chan}, \bibinfo{person}{Tao Chen}, \bibinfo{person}{Abhishek Gupta}, {and} \bibinfo{person}{Pulkit Agrawal}.} \bibinfo{year}{2024}\natexlab{}.
\newblock \showarticletitle{Reconciling reality through simulation: A real-to-sim-to-real approach for robust manipulation}.
\newblock \bibinfo{journal}{\emph{arXiv preprint arXiv:2403.03949}} (\bibinfo{year}{2024}).
\newblock


\bibitem[Umeyama(1991)]%
        {umeyama1991least}
\bibfield{author}{\bibinfo{person}{Shinji Umeyama}.} \bibinfo{year}{1991}\natexlab{}.
\newblock \showarticletitle{Least-squares estimation of transformation parameters between two point patterns}.
\newblock \bibinfo{journal}{\emph{IEEE Transactions on Pattern Analysis \& Machine Intelligence}} \bibinfo{volume}{13}, \bibinfo{number}{04} (\bibinfo{year}{1991}), \bibinfo{pages}{376--380}.
\newblock


\bibitem[Voleti et~al\mbox{.}(2025)]%
        {voleti2025sv3d}
\bibfield{author}{\bibinfo{person}{Vikram Voleti}, \bibinfo{person}{Chun-Han Yao}, \bibinfo{person}{Mark Boss}, \bibinfo{person}{Adam Letts}, \bibinfo{person}{David Pankratz}, \bibinfo{person}{Dmitry Tochilkin}, \bibinfo{person}{Christian Laforte}, \bibinfo{person}{Robin Rombach}, {and} \bibinfo{person}{Varun Jampani}.} \bibinfo{year}{2025}\natexlab{}.
\newblock \showarticletitle{Sv3d: Novel multi-view synthesis and 3d generation from a single image using latent video diffusion}. In \bibinfo{booktitle}{\emph{European Conference on Computer Vision}}. Springer, \bibinfo{pages}{439--457}.
\newblock


\bibitem[Wang et~al\mbox{.}(2024b)]%
        {wang2024moge}
\bibfield{author}{\bibinfo{person}{Ruicheng Wang}, \bibinfo{person}{Sicheng Xu}, \bibinfo{person}{Cassie Dai}, \bibinfo{person}{Jianfeng Xiang}, \bibinfo{person}{Yu Deng}, \bibinfo{person}{Xin Tong}, {and} \bibinfo{person}{Jiaolong Yang}.} \bibinfo{year}{2024}\natexlab{b}.
\newblock \showarticletitle{Moge: Unlocking accurate monocular geometry estimation for open-domain images with optimal training supervision}.
\newblock \bibinfo{journal}{\emph{arXiv preprint arXiv:2410.19115}} (\bibinfo{year}{2024}).
\newblock


\bibitem[Wang et~al\mbox{.}(2024a)]%
        {wang2024prolificdreamer}
\bibfield{author}{\bibinfo{person}{Zhengyi Wang}, \bibinfo{person}{Cheng Lu}, \bibinfo{person}{Yikai Wang}, \bibinfo{person}{Fan Bao}, \bibinfo{person}{Chongxuan Li}, \bibinfo{person}{Hang Su}, {and} \bibinfo{person}{Jun Zhu}.} \bibinfo{year}{2024}\natexlab{a}.
\newblock \showarticletitle{Prolificdreamer: High-fidelity and diverse text-to-3d generation with variational score distillation}.
\newblock \bibinfo{journal}{\emph{Advances in Neural Information Processing Systems}}  \bibinfo{volume}{36} (\bibinfo{year}{2024}).
\newblock


\bibitem[Wei et~al\mbox{.}(2022)]%
        {wei2022approximate}
\bibfield{author}{\bibinfo{person}{Xinyue Wei}, \bibinfo{person}{Minghua Liu}, \bibinfo{person}{Zhan Ling}, {and} \bibinfo{person}{Hao Su}.} \bibinfo{year}{2022}\natexlab{}.
\newblock \showarticletitle{Approximate convex decomposition for 3d meshes with collision-aware concavity and tree search}.
\newblock \bibinfo{journal}{\emph{ACM Transactions on Graphics (TOG)}} \bibinfo{volume}{41}, \bibinfo{number}{4} (\bibinfo{year}{2022}), \bibinfo{pages}{1--18}.
\newblock


\bibitem[Wu et~al\mbox{.}(2024a)]%
        {wu2024unique3d}
\bibfield{author}{\bibinfo{person}{Kailu Wu}, \bibinfo{person}{Fangfu Liu}, \bibinfo{person}{Zhihan Cai}, \bibinfo{person}{Runjie Yan}, \bibinfo{person}{Hanyang Wang}, \bibinfo{person}{Yating Hu}, \bibinfo{person}{Yueqi Duan}, {and} \bibinfo{person}{Kaisheng Ma}.} \bibinfo{year}{2024}\natexlab{a}.
\newblock \showarticletitle{Unique3D: High-Quality and Efficient 3D Mesh Generation from a Single Image}.
\newblock \bibinfo{journal}{\emph{arXiv preprint arXiv:2405.20343}} (\bibinfo{year}{2024}).
\newblock


\bibitem[Wu et~al\mbox{.}(2024b)]%
        {wu2024reconfusion}
\bibfield{author}{\bibinfo{person}{Rundi Wu}, \bibinfo{person}{Ben Mildenhall}, \bibinfo{person}{Philipp Henzler}, \bibinfo{person}{Keunhong Park}, \bibinfo{person}{Ruiqi Gao}, \bibinfo{person}{Daniel Watson}, \bibinfo{person}{Pratul~P Srinivasan}, \bibinfo{person}{Dor Verbin}, \bibinfo{person}{Jonathan~T Barron}, \bibinfo{person}{Ben Poole}, {et~al\mbox{.}}} \bibinfo{year}{2024}\natexlab{b}.
\newblock \showarticletitle{Reconfusion: 3d reconstruction with diffusion priors}. In \bibinfo{booktitle}{\emph{Proceedings of the IEEE/CVF Conference on Computer Vision and Pattern Recognition}}. \bibinfo{pages}{21551--21561}.
\newblock


\bibitem[Xiang et~al\mbox{.}(2024)]%
        {xiang2024structured}
\bibfield{author}{\bibinfo{person}{Jianfeng Xiang}, \bibinfo{person}{Zelong Lv}, \bibinfo{person}{Sicheng Xu}, \bibinfo{person}{Yu Deng}, \bibinfo{person}{Ruicheng Wang}, \bibinfo{person}{Bowen Zhang}, \bibinfo{person}{Dong Chen}, \bibinfo{person}{Xin Tong}, {and} \bibinfo{person}{Jiaolong Yang}.} \bibinfo{year}{2024}\natexlab{}.
\newblock \showarticletitle{Structured 3D Latents for Scalable and Versatile 3D Generation}.
\newblock \bibinfo{journal}{\emph{arXiv preprint arXiv:2412.01506}} (\bibinfo{year}{2024}).
\newblock


\bibitem[Xiao et~al\mbox{.}(2024)]%
        {xiao2024florence}
\bibfield{author}{\bibinfo{person}{Bin Xiao}, \bibinfo{person}{Haiping Wu}, \bibinfo{person}{Weijian Xu}, \bibinfo{person}{Xiyang Dai}, \bibinfo{person}{Houdong Hu}, \bibinfo{person}{Yumao Lu}, \bibinfo{person}{Michael Zeng}, \bibinfo{person}{Ce Liu}, {and} \bibinfo{person}{Lu Yuan}.} \bibinfo{year}{2024}\natexlab{}.
\newblock \showarticletitle{Florence-2: Advancing a unified representation for a variety of vision tasks}. In \bibinfo{booktitle}{\emph{Proceedings of the IEEE/CVF Conference on Computer Vision and Pattern Recognition}}. \bibinfo{pages}{4818--4829}.
\newblock


\bibitem[Xie et~al\mbox{.}(2024)]%
        {xie2024physgaussian}
\bibfield{author}{\bibinfo{person}{Tianyi Xie}, \bibinfo{person}{Zeshun Zong}, \bibinfo{person}{Yuxing Qiu}, \bibinfo{person}{Xuan Li}, \bibinfo{person}{Yutao Feng}, \bibinfo{person}{Yin Yang}, {and} \bibinfo{person}{Chenfanfu Jiang}.} \bibinfo{year}{2024}\natexlab{}.
\newblock \showarticletitle{Physgaussian: Physics-integrated 3d gaussians for generative dynamics}. In \bibinfo{booktitle}{\emph{Proceedings of the IEEE/CVF Conference on Computer Vision and Pattern Recognition}}. \bibinfo{pages}{4389--4398}.
\newblock


\bibitem[Xu et~al\mbox{.}(2024)]%
        {xu2024precise}
\bibfield{author}{\bibinfo{person}{Qingshan Xu}, \bibinfo{person}{Jiao Liu}, \bibinfo{person}{Melvin Wong}, \bibinfo{person}{Caishun Chen}, {and} \bibinfo{person}{Yew-Soon Ong}.} \bibinfo{year}{2024}\natexlab{}.
\newblock \showarticletitle{Precise-Physics Driven Text-to-3D Generation}.
\newblock \bibinfo{journal}{\emph{arXiv preprint arXiv:2403.12438}} (\bibinfo{year}{2024}).
\newblock


\bibitem[Yang et~al\mbox{.}(2023)]%
        {yang2023setofmark}
\bibfield{author}{\bibinfo{person}{Jianwei Yang}, \bibinfo{person}{Hao Zhang}, \bibinfo{person}{Feng Li}, \bibinfo{person}{Xueyan Zou}, \bibinfo{person}{Chunyuan Li}, {and} \bibinfo{person}{Jianfeng Gao}.} \bibinfo{year}{2023}\natexlab{}.
\newblock \showarticletitle{Set-of-Mark Prompting Unleashes Extraordinary Visual Grounding in GPT-4V}.
\newblock \bibinfo{journal}{\emph{arXiv preprint arXiv:2310.11441}} (\bibinfo{year}{2023}).
\newblock


\bibitem[Yang et~al\mbox{.}(2024b)]%
        {yang2024depth}
\bibfield{author}{\bibinfo{person}{Lihe Yang}, \bibinfo{person}{Bingyi Kang}, \bibinfo{person}{Zilong Huang}, \bibinfo{person}{Xiaogang Xu}, \bibinfo{person}{Jiashi Feng}, {and} \bibinfo{person}{Hengshuang Zhao}.} \bibinfo{year}{2024}\natexlab{b}.
\newblock \showarticletitle{Depth anything: Unleashing the power of large-scale unlabeled data}. In \bibinfo{booktitle}{\emph{Proceedings of the IEEE/CVF Conference on Computer Vision and Pattern Recognition}}. \bibinfo{pages}{10371--10381}.
\newblock


\bibitem[Yang et~al\mbox{.}(2024a)]%
        {yang2024physcene}
\bibfield{author}{\bibinfo{person}{Yandan Yang}, \bibinfo{person}{Baoxiong Jia}, \bibinfo{person}{Peiyuan Zhi}, {and} \bibinfo{person}{Siyuan Huang}.} \bibinfo{year}{2024}\natexlab{a}.
\newblock \showarticletitle{Physcene: Physically interactable 3d scene synthesis for embodied ai}. In \bibinfo{booktitle}{\emph{Proceedings of the IEEE/CVF Conference on Computer Vision and Pattern Recognition}}. \bibinfo{pages}{16262--16272}.
\newblock


\bibitem[Yin et~al\mbox{.}(2023)]%
        {yin2023metric3d}
\bibfield{author}{\bibinfo{person}{Wei Yin}, \bibinfo{person}{Chi Zhang}, \bibinfo{person}{Hao Chen}, \bibinfo{person}{Zhipeng Cai}, \bibinfo{person}{Gang Yu}, \bibinfo{person}{Kaixuan Wang}, \bibinfo{person}{Xiaozhi Chen}, {and} \bibinfo{person}{Chunhua Shen}.} \bibinfo{year}{2023}\natexlab{}.
\newblock \showarticletitle{Metric3d: Towards zero-shot metric 3d prediction from a single image}. In \bibinfo{booktitle}{\emph{Proceedings of the IEEE/CVF International Conference on Computer Vision}}. \bibinfo{pages}{9043--9053}.
\newblock


\bibitem[Yu et~al\mbox{.}(2021)]%
        {yu2021pixelnerf}
\bibfield{author}{\bibinfo{person}{Alex Yu}, \bibinfo{person}{Vickie Ye}, \bibinfo{person}{Matthew Tancik}, {and} \bibinfo{person}{Angjoo Kanazawa}.} \bibinfo{year}{2021}\natexlab{}.
\newblock \showarticletitle{pixelnerf: Neural radiance fields from one or few images}. In \bibinfo{booktitle}{\emph{Proceedings of the IEEE/CVF conference on computer vision and pattern recognition}}. \bibinfo{pages}{4578--4587}.
\newblock


\bibitem[Yu et~al\mbox{.}(2024)]%
        {yu2024wonderjourney}
\bibfield{author}{\bibinfo{person}{Hong-Xing Yu}, \bibinfo{person}{Haoyi Duan}, \bibinfo{person}{Junhwa Hur}, \bibinfo{person}{Kyle Sargent}, \bibinfo{person}{Michael Rubinstein}, \bibinfo{person}{William~T Freeman}, \bibinfo{person}{Forrester Cole}, \bibinfo{person}{Deqing Sun}, \bibinfo{person}{Noah Snavely}, \bibinfo{person}{Jiajun Wu}, {et~al\mbox{.}}} \bibinfo{year}{2024}\natexlab{}.
\newblock \showarticletitle{Wonderjourney: Going from anywhere to everywhere}. In \bibinfo{booktitle}{\emph{Proceedings of the IEEE/CVF Conference on Computer Vision and Pattern Recognition}}. \bibinfo{pages}{6658--6667}.
\newblock


\bibitem[Yu et~al\mbox{.}(2022)]%
        {yu2022monosdf}
\bibfield{author}{\bibinfo{person}{Zehao Yu}, \bibinfo{person}{Songyou Peng}, \bibinfo{person}{Michael Niemeyer}, \bibinfo{person}{Torsten Sattler}, {and} \bibinfo{person}{Andreas Geiger}.} \bibinfo{year}{2022}\natexlab{}.
\newblock \showarticletitle{Monosdf: Exploring monocular geometric cues for neural implicit surface reconstruction}.
\newblock \bibinfo{journal}{\emph{Advances in neural information processing systems}}  \bibinfo{volume}{35} (\bibinfo{year}{2022}), \bibinfo{pages}{25018--25032}.
\newblock


\bibitem[Yuille and Kersten(2006)]%
        {yuille2006vision}
\bibfield{author}{\bibinfo{person}{Alan Yuille} {and} \bibinfo{person}{Daniel Kersten}.} \bibinfo{year}{2006}\natexlab{}.
\newblock \showarticletitle{Vision as Bayesian inference: analysis by synthesis?}
\newblock \bibinfo{journal}{\emph{Trends in cognitive sciences}} \bibinfo{volume}{10}, \bibinfo{number}{7} (\bibinfo{year}{2006}), \bibinfo{pages}{301--308}.
\newblock


\bibitem[Zhang et~al\mbox{.}(2023)]%
        {zhang20233dshape2vecset}
\bibfield{author}{\bibinfo{person}{Biao Zhang}, \bibinfo{person}{Jiapeng Tang}, \bibinfo{person}{Matthias Niessner}, {and} \bibinfo{person}{Peter Wonka}.} \bibinfo{year}{2023}\natexlab{}.
\newblock \showarticletitle{3dshape2vecset: A 3d shape representation for neural fields and generative diffusion models}.
\newblock \bibinfo{journal}{\emph{ACM Transactions on Graphics (TOG)}} \bibinfo{volume}{42}, \bibinfo{number}{4} (\bibinfo{year}{2023}), \bibinfo{pages}{1--16}.
\newblock


\bibitem[Zhang et~al\mbox{.}(2024b)]%
        {zhang2024improving}
\bibfield{author}{\bibinfo{person}{Jiancheng Zhang}, \bibinfo{person}{Haijin Zeng}, \bibinfo{person}{Yongyong Chen}, \bibinfo{person}{Dengxiu Yu}, {and} \bibinfo{person}{Yin-Ping Zhao}.} \bibinfo{year}{2024}\natexlab{b}.
\newblock \showarticletitle{Improving Spectral Snapshot Reconstruction with Spectral-Spatial Rectification}. In \bibinfo{booktitle}{\emph{Proceedings of the IEEE/CVF Conference on Computer Vision and Pattern Recognition}}. \bibinfo{pages}{25817--25826}.
\newblock


\bibitem[Zhang et~al\mbox{.}(2024a)]%
        {zhang2024clay}
\bibfield{author}{\bibinfo{person}{Longwen Zhang}, \bibinfo{person}{Ziyu Wang}, \bibinfo{person}{Qixuan Zhang}, \bibinfo{person}{Qiwei Qiu}, \bibinfo{person}{Anqi Pang}, \bibinfo{person}{Haoran Jiang}, \bibinfo{person}{Wei Yang}, \bibinfo{person}{Lan Xu}, {and} \bibinfo{person}{Jingyi Yu}.} \bibinfo{year}{2024}\natexlab{a}.
\newblock \showarticletitle{CLAY: A Controllable Large-scale Generative Model for Creating High-quality 3D Assets}.
\newblock \bibinfo{journal}{\emph{ACM Transactions on Graphics (TOG)}} \bibinfo{volume}{43}, \bibinfo{number}{4} (\bibinfo{year}{2024}), \bibinfo{pages}{1--20}.
\newblock


\bibitem[Zhengwentai(2023)]%
        {taited2023CLIPScore}
\bibfield{author}{\bibinfo{person}{SUN Zhengwentai}.} \bibinfo{year}{2023}\natexlab{}.
\newblock \bibinfo{title}{{clip-score: CLIP Score for PyTorch}}.
\newblock \bibinfo{howpublished}{\url{https://github.com/taited/clip-score}}.
\newblock
\newblock
\shownote{Version 0.1.1}.


\bibitem[Zhong et~al\mbox{.}(2025)]%
        {zhong2025reconstruction}
\bibfield{author}{\bibinfo{person}{Licheng Zhong}, \bibinfo{person}{Hong-Xing Yu}, \bibinfo{person}{Jiajun Wu}, {and} \bibinfo{person}{Yunzhu Li}.} \bibinfo{year}{2025}\natexlab{}.
\newblock \showarticletitle{Reconstruction and simulation of elastic objects with spring-mass 3d gaussians}. In \bibinfo{booktitle}{\emph{European Conference on Computer Vision}}. Springer, \bibinfo{pages}{407--423}.
\newblock


\bibitem[Zhou et~al\mbox{.}(2018)]%
        {Zhou2018Open3DAM}
\bibfield{author}{\bibinfo{person}{Qian-Yi Zhou}, \bibinfo{person}{Jaesik Park}, {and} \bibinfo{person}{Vladlen Koltun}.} \bibinfo{year}{2018}\natexlab{}.
\newblock \showarticletitle{Open3D: A Modern Library for 3D Data Processing}.
\newblock \bibinfo{journal}{\emph{ArXiv}}  \bibinfo{volume}{abs/1801.09847} (\bibinfo{year}{2018}).
\newblock


\end{thebibliography}
